\documentclass[10pt,twocolumn,letterpaper]{article}

\usepackage[pagenumbers]{cvpr} 

\usepackage[dvipsnames]{xcolor}

\usepackage[accsupp]{axessibility}

\definecolor{gary}{RGB}{0,120,120}

\usepackage{float}
\usepackage{booktabs}
\usepackage{multirow}
\usepackage{xcolor,colortbl}
\usepackage{pifont}
\usepackage[dvipsnames]{xcolor}

\usepackage[normalem]{ulem}
\useunder{\uline}{\ul}{}
\definecolor{cvprblue}{rgb}{0.21,0.49,0.74}
\usepackage[pagebackref,breaklinks,colorlinks,allcolors=cvprblue]{hyperref}

\newcommand\blfootnote[1]{%
  \begingroup
  \renewcommand\thefootnote{}\footnote{#1}%
  \addtocounter{footnote}{-1}%
  \endgroup
}

\title{InsTaG: Learning Personalized 3D Talking Head from Few-Second Video}

\author{
    Jiahe Li$^1$ \quad Jiawei Zhang$^1$ \quad Xiao Bai$^1$\thanks{Corresponding author: Xiao Bai (baixiao@buaa.edu.cn).} \quad Jin Zheng$^1$ \quad Jun Zhou$^2$ \quad Lin Gu$^{3,4}$ \\
    $^1$School of Computer Science and Engineering, State\,Key\,Laboratory\,of\,Complex\,\&\,Critical\,Software \\ Environment,\, Jiangxi Research Institute,\, Beihang University\\
    $^2$School of Information and Communication Technology,\, Griffith University\\
    $^3$RIKEN AIP \quad $^4$The University of Tokyo
} 

\begin{document}

\twocolumn[{
\renewcommand\twocolumn[1][]{#1}
\maketitle

\begin{center} 
    \vspace{-5mm}
    \setlength{\abovecaptionskip}{12pt}
    \centering
    \includegraphics[width=1.0\linewidth]{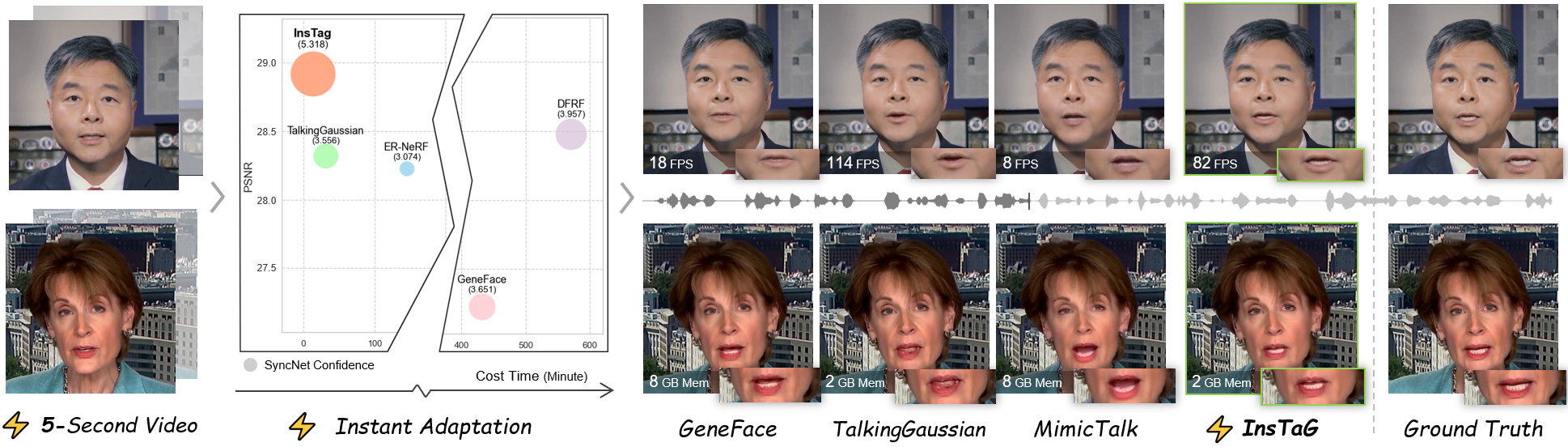}
    \captionof{figure}{With only 5-second video data, InsTaG outperforms the state-of-the-arts \cite{ye2023geneface, li2024talkinggaussian, ye2024mimictalk} by delivering high-quality personalized lip synchronization and realistic rendering with the fastest adaptation, meanwhile attaining low memory overhead and real-time inference. } 
    \label{fig:demo1} 
\end{center}
}
]

\begin{abstract}
Despite exhibiting impressive performance in synthesizing lifelike personalized 3D talking heads, prevailing methods based on radiance fields suffer from high demands for training data and time for each new identity. This paper introduces InsTaG, a 3D talking head synthesis framework that allows a fast learning of realistic personalized 3D talking head from few training data. Built upon a lightweight 3DGS person-specific synthesizer with universal motion priors, InsTaG achieves high-quality and fast adaptation while preserving high-level personalization and efficiency. As preparation, we first propose an Identity-Free Pre-training strategy that enables the pre-training of the person-specific model and encourages the collection of universal motion priors from long-video data corpus. To fully exploit the universal motion priors to learn an unseen new identity, we then present a Motion-Aligned Adaptation strategy to adaptively align the target head to the pre-trained field, and constrain a robust dynamic head structure under few training data. Experiments demonstrate our outstanding performance and efficiency under various data scenarios to render high-quality personalized talking heads. Project page: \url{https://fictionarry.github.io/InsTaG/} .

\end{abstract}

\vspace{-4mm}
\section{Introduction}
Audio-driven talking head synthesis has become an important technique for various digital applications like video production, virtual reality, and human-computer interaction.
The recent advancements in Neural Radiance Fields (NeRF) \cite{mildenhall2021nerf} and 3D Gaussian Splatting (3DGS) \cite{kerbl2023gaussian} have led to significant progress in 3D representations. Various methods \cite{guo2021ad, shen2022dfrf, liu2022semantic, ye2023geneface, li2024er, li2024talkinggaussian} now utilize radiance fields to create high-fidelity 3D talking head videos, capturing personalized talking styles with impressive photorealistic image quality.
\blfootnote{$^*$Corresponding author: Xiao Bai (baixiao@buaa.edu.cn).}

Most of these methods adopt person-specific training. Given a video clip with the target talking person, a whole person-specific model is trained to memorize the target talking head. Despite their tremendous improvements in high-fidelity rendering, personalized dynamics, and fast running speed, as a trade-off, the model holds no generalizability for an unseen identity. Thus, they require a large amount of high-quality video frames with a long training time to adapt to each new identity. Some image-referred methods \cite{ye2024real3d, li2023hide, shen2022dfrf, ye2024mimictalk} can quickly generalize to different identities, nevertheless, the quality of personalization and efficiency drop. A question is then raised: \emph{How can we use the least training data to rapidly learn a faithful 3D talking head, while preserving a high level of personalization?}

Observing previous radiance-field-based methods, we notice that: \textbf{1)} While the training process includes a long time to learn a basic audio-driven motion that with only little personalized style, the static 3D head can be well reconstructed in just a few early iterations; \textbf{2)} Though the results may exhibit jittering and poor quality when trained with limited data, a portion of the lip motions remains recognizable with distinct personalized style. These suggest that even a short video clip can provide rich personalization cues to enable the learning of a specific talking head with little time cost, as long as common audio-motion knowledge is obtained. Motivated by this, by preparing a pre-trained identity-free 3D motion field as prior and aligning it to the new identity to facilitate the adaptation, we propose Instant Talking Head Synthesis with Gaussian Splatting (\textbf{InsTaG}), a framework that allows fast learning of high-fidelity personalized 3D talking head from a video even as short as only a few seconds while attaining high efficiency.  

Different from previous few-shot methods \cite{ye2024real3d, li2023hide, ye2024mimictalk} that imitate a new identity from inputs, InsTaG memorizes the whole talking head in a lightweight person-specific model to obtain high-level personalization and efficiency. Upon such a person-specific paradigm, we fully decouple the learning of universal motion and personalization into pre-training and adaptation, preparing a pre-trained 3D motion field for future person-specific learning of new identity. However, a challenge with this design lies in pre-training with multi-person data, where the person-specific model can not fit varying identities and estimate their varied personalized motions, leading to an ineffective optimization. To tackle this problem, we present an \textit{Identity-Free Pre-training} strategy. By keeping a series of temporary personal fields to store the identity information and filter out the personalized motions, we extract the common knowledge of talking motion from long-video corpus and produce a universal motion field that contains identity-free motion prior. To further facilitate this motion detachment, a negative contrast loss is introduced to encourage the diversity of learned personalized motion between each sample, maximizing the retaining of universal priors for later adaptation.

To fully exploit the universal motion priors to enable fast and high-quality new-identity adaptation, we present a \textit{Motion-Aligned Adaptation} strategy to align the unseen talking head to the pre-trained motion field. First, a motion aligner is developed to learn a condition-independent primitive-wise coordinate offset and a motion scaler. It helps the intra-alignment between the personalized and universal motions in the field, maximizing the retention of learned knowledge during few-shot fine-tuning. Moreover, to enhance the generalizability by face-mouth decomposition \cite{yu2022talkingdiff} while preserving a robust dynamic face structure, we introduce a face-mouth hook technique to inter-align the inside mouth motion with the face, which efficiently improves the fidelity as well as motion quality. To further improve the robustness of view direction, the new identity's head structure is aligned with estimated geometry priors for regularization. By incorporating the alignments in the adaptation, the person-specific model finally synthesizes high-fidelity personalized 3D talking heads at a fast speed.

Integrating the two proposed strategies, InsTaG can learn a personalized 3D talking head from even a short 5-second video, achieving state-of-the-art visual quality and personalization compared to existing methods, with high lip-synchronization and real-time inference. Experiments demonstrate the outperforming efficiency and generalizability of InsTaG for various identities, genders and languages.

Our main contributions are summarized as follows:

\begin{itemize}
    \item An \textit{Identity-Free Pre-training} strategy that enables the pretraining of the person-specific model by filtering out conflict identity and personalized motions, meanwhile maximizing the capturing of universal motion priors.
    \item A \textit{Motion-Aligned Adaptation} strategy, which inter- and intra-aligns the talking head with the motion field for efficient new-identity adaptation, attaining robust, realistic reconstruction and personalized lip-synchronization.
    \item Extensive experiments show that our InsTaG learns realistic personalized talking heads from few data, while attaining high lip-synchronization and efficiency, outperforming state-of-the-art methods under various scenarios.
\end{itemize}

\begin{figure*}[!t]
    \centering
    \setlength{\abovecaptionskip}{7pt}
    \includegraphics[width=1\linewidth]{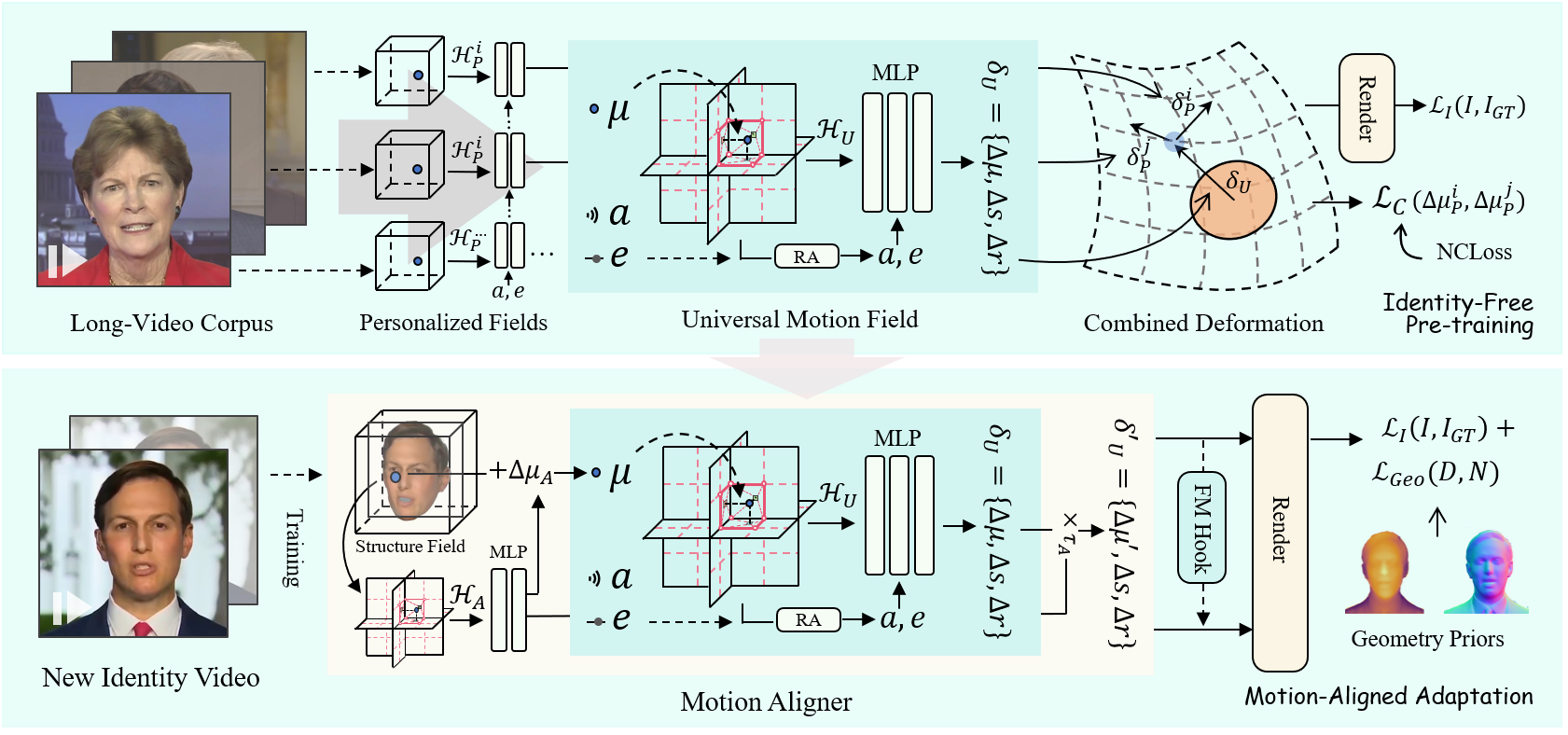}
    \caption{\textbf{Overview of InsTaG.} For preparation, InsTaG collects the common knowledge of talking motion from a long-video corpus by \textit{Identity-Free Pre-training}, storing it as a motion field. Given a short video with a new identity, the \textit{Motion-Aligned Adaptation} strategy builds a robust and fast person-specific synthesizer with the pre-trained motion field to learn a high-quality personalized 3D talking head.}
    \label{fig:main}
    \vspace{-3.5mm}
\end{figure*}

\vspace{-1mm}
\section{Related Work}
\vspace{-1mm}

\noindent\textbf{3D Talking Head Synthesis.}
Talking head synthesis aims to use arbitrary audio to reenact a talking person to generate audio-visual synchronized videos. Early works \cite{prajwal2020wav2lip, ezzat2002trainable, jamaludin2019you, chen2019hierarchical, wiles2018x2face, zhou2021pcavs, wang2020mead} are mainly built on generative models, creating talking heads by manipulating the given 2D images. Later, to solve the temporal inconsistency when the head moves, 3D-based methods \cite{thies2020nvp, yi2020audio, zhang2021facial, lu2021lsp} utilize explicit 3D face structure and successfully improve the naturalness. 

Recently, radiance fields like NeRF~\cite{mildenhall2021nerf} and 3D Gaussian Splatting (3DGS) \cite{kerbl2023gaussian} have been introduced as the representation to allow 3D talking head synthesize. Inheriting the radiance field optimization, most methods \cite{guo2021ad, liu2022semantic, tang2022rad, li2023efficient, peng2023synctalk, li2024talkinggaussian, cho2024gaussiantalker} train a person-specific model on minutes of high-quality video, achieving success in reconstructing photorealistic rendering and personalized talking style. However, the strict quality requirements of the training video data and the long training time for every adaptation to a new identity have hugely limited their application.
Although some works \cite{shen2022dfrf, ye2023geneface, li2023hide, yu2024gaussiantalker, ye2024real3d, ye2024mimictalk} attempt to solve the problem by a one-shot generator \cite{li2023hide, ye2024real3d, ye2024mimictalk} or take pre-trained motions from external modules to loosely join the adaptation \cite{shen2022dfrf, ye2023geneface} to achieve one- or few-shot, they often compromise image quality and loses the personalization as the trade-off. A concurrent work MimicTalk \cite{ye2024mimictalk} injects LoRA \cite{hu2021lora} to improve fine-tuning efficiency, but its inference is slow due to the large backbone.
Instead, by enabling the pretraining of person-specific models and using pre-trained motion field to directly drive the talking head, our method obtains a more compact and consistent architecture to fully exploit the motion priors in an end-to-end way, facilitating modeling precise lip-synchronization with high efficiency. 

\noindent\textbf{Few-shot 3D Head Reconstruction.}
Fitting high-fidelity 3D human heads from RGB images has been a hot topic. Early model-based methods \cite{yu2018headfusion, thies2016face2face, yang2020facescape, gecer2019ganfit} focus on fitting a 3D head with 3DMMs \cite{paysan2009bfm, FLAME:SiggraphAsia2017, booth20173d} but are short in detail. Combined with neural fields, some model-free methods \cite{buhler2023preface, ramon2021h3d, mihajlovic2022keypointnerf} learn static heads from a few images with structure priors. Although many works  \cite{giebenhain2024mononphm, gafni2021dynamic, song2024tri, xu2023avatarmav, yang2024have, hong2022headnerf, cao2022authentic} attempt to reconstruct dynamic heads from few videos, their parametric controls are difficult to predict from audio and limit the mouth motion granularity, making it hard for them to express the highly personalized 3D talking head with delicate audio-driven mouth motions achieved by InsTaG. 

Some generative methods  \cite{zakharov2019few, gan2023eat, xu2024vasa, gururani2023space, li2023hide, ye2024real3d} focus on reconstructing talking heads with 3D structure, however, they can hardly infer personalized talking style from only one image, leading to weak personalization and fidelity. Learning 3D talking heads with few-shot NeRF, recent works \cite{shen2022dfrf, ye2024mimictalk, li2024ae} introduce 2D-to-3D modules with pre-training but are often slow due to their overhead. 
Instead, with an identity-free motion prior and person-specific adaptation instead of expensive 2D-to-3D priors, our InsTaG can rapidly learn a personalized audio-driven 3D talking head from few data while attaining fast inference.

\vspace{-1mm}
\section{InsTaG}
\vspace{-1mm}

As shown in Figure \ref{fig:main}, InsTaG synthesizes personalized 3D talking head based on Gaussian Splatting \cite{kerbl2023gaussian} (Sec. \ref{sec:synthesizer}), consisting of an Identity-Free Pre-training  (Sec. \ref{sec:pretrain}) strategy to fetch universal motion priors and a Motion-Aligned Adaptation strategy (Sec. \ref{sec:adaptation}) to learn a new identity. 
We describe the detailed designs in the following sections.

 \vspace{-1mm}
\subsection{3DGS Talking Head Synthesizer \label{sec:synthesizer}}
\vspace{-1mm}

Our framework aims to learn personalized 3D talking head with fast inference and instant adaptation, therefore, a lightweight model with high representation ability is needed to be as the 3D synthesizer. Generally, existing methods can be categorized into image-based \cite{ye2024real3d, li2023hide, ye2024mimictalk} and person-specific \cite{guo2021ad, tang2022rad, li2023efficient, li2024talkinggaussian}. The former often possesses generalizability but is relatively slow and weak in personalization, while the latter matches our requirements. 

\noindent\textbf{Preliminaries.}
A person-specific synthesizer based on radiance fields can be formulated as a mapping $\mathcal{F}$ that from given conditions to a synthesized head image $I$:
\begin{equation}
    \label{eq:problem}
    \setlength{\abovedisplayskip}{6pt}
    \setlength{\belowdisplayskip}{6pt}
    \mathcal{F}: (\boldsymbol{a}, \boldsymbol{e}, [R, t]; \theta) \rightarrow I
\end{equation}
where the conditions include audio $\boldsymbol{a}$, upper-face expression control $\boldsymbol{e}$ and camera pose $[R, t]$ that represents the head pose, and the parameters of the radiance fields $\theta$.  

Under such an architecture, we require the motion learning to be entirely irrelevant to appearance for pre-training. Hence, we leverage the recent 3DGS-based synthesizer with a face-mouth decomposition \cite{li2024talkinggaussian}, including a pair of structure fields and motion fields separately for a face branch and inside-mouth branch to synthesize the talking head image $I$. 

\noindent\textbf{Structure Field.}
The structure field stores the static Gaussian primitives with parameters $\theta = \{\mu, s, q, \alpha, f\}$. Both the face branch and inside-mouth branch obtain a private structure field for their own target structure.

\noindent\textbf{Motion Field.}
The motion field predicts a point-wise deformation $\delta = \{\Delta \mu, \Delta s, \Delta q\}$ with a tri-plane hash encoder $\mathcal{H}$ \cite{li2023efficient} with a region attention (RA) module to store spatial relation, and an MLP decoder to predict deformation. For each query primitive with center $\mu$, the neural field $\mathcal{D}$ predicts its deformation $\delta$ from the given condition feature set $\mathbf{C}$ including audio $\boldsymbol{a}$ and upper-face expression signal $\boldsymbol{e}$:
\begin{equation}
    \label{eq:motionfield}
    \setlength{\abovedisplayskip}{6pt}
    \setlength{\belowdisplayskip}{6pt}
    \delta = \mathcal{D}(\mu, \mathbf{C}) = \text{MLP}(\mathcal{H}(\mu) \oplus \mathbf{C}),
\end{equation}
where $\oplus$ denotes concatenation.

\noindent\textbf{Rendering.}
During rendering, the motion field deforms the Gaussians $\theta$ depending on the inputs $\boldsymbol{a}$ and $\boldsymbol{e}$, and then a 2D image is rendered. 
This process happens in both the face and inside-mouth branches, and the output results are combined to get the final talking head image.

In the following, unless necessary, we do not show the face-mouth decomposition for easier understanding.

\vspace{-1mm}
\subsection{Identity-Free Pre-training \label{sec:pretrain}}
\vspace{-1mm}
Although with personalization knowledge, a short video clip is insufficient to train a talking head with personalized audio-motion mapping from scratch. To this end, we extract the common motion knowledge from long videos by pre-training as compensation for the new identity.

In contrast to existing generative pre-training solutions \cite{shen2022dfrf, li2023hide, ye2024mimictalk} that allow multiple-identity inputs to train the same model, an identify-conflict problem lies in our method. Firstly, different identities can not coexist in one person-specific model. Secondly, since the model can not separate different people, the conflicting personalized motion hinders the model convergence and pollutes the pre-trained module by overfitting the noises.  

\noindent\textbf{Strategy.} To tackle this problem, we design storing the identity-dependent knowledge in a \textit{Personalized Field} distributed to each training video, therefore detaching the identity influence of the learning of universal motion to pre-train a \textit{Universal Motion Field}. Specifically, each Personalized Field consists of a personalized structure field with parameter $\theta_P$ and a smaller motion field $\mathcal{D}_P$ to retain the identity appearance and personalized motion of its owner video. 

\noindent\textbf{Universal Motion Field (UMF).} Serving as a pre-trained module, UMF aims to predict a commonly correct facial motion that suits most identities. Queried by an arbitrary Gaussian, the field $\mathcal{U}$ predicts the universal deformation $\delta_U$ without personalization according to Eq. (\ref{eq:motionfield}):
\begin{equation}
    \setlength{\abovedisplayskip}{6pt}
    \setlength{\belowdisplayskip}{6pt}
    \delta_U = \mathcal{U}(\mu, \mathbf{C}) = \text{MLP}(\mathcal{H}_U(\mu) \oplus \mathbf{C}).
\end{equation}

\noindent\textbf{Personalized Fields. } During training, Personalized Fields cooperate with one sharing Universal Motion Field to synthesize personalized talking heads, enabling the supervision from videos with different identities. 
Practically, given an audio feature $\boldsymbol{a}$ and expression signal $\boldsymbol{e}$ packaged by $\mathbf{C}$, the UMF $\mathcal{U}$ would predict an identity-free deformation queried by $\mu_P^i \in \theta_P^i$, serving as a basic deformation for the $i$-th person. Meanwhile, the $i$-th personalized field predicts a residual motion $\delta_P^i$ as the personalized adjustment:
\vspace{-1mm}
\begin{equation}
    \label{eq:pdeform}
    \setlength{\abovedisplayskip}{6pt}
    \setlength{\belowdisplayskip}{6pt}
    \delta_P^i = \mathcal{D}_P^i(\mu_P^i, \mathbf{C}) = \text{MLP}(\mathcal{H}_P^i(\mu_P^i) \oplus \mathbf{C}).
\end{equation}
After that, the two deformations are combined to deform the personalized structure fields for rendering:
\begin{equation}
    \setlength{\abovedisplayskip}{6pt}
    \setlength{\belowdisplayskip}{6pt}
    \mathcal{R}: ([R, t], \widetilde{\theta}_P^i) \rightarrow I, \quad \widetilde{\theta}_P^i = \theta_P^i + \delta_U + \delta_P^i,
\end{equation}
where $\mathcal{R}$ refers to the 3DGS rendering process.

Since then, videos with different talking identities can be used for pre-training with no identity conflicts, and the personalized motions are isolated from each other by staying at the personalized fields, detached from the universal ones.

\noindent\textbf{Negative Contrast Loss.}
Despite the detachment of personalized and universal motions enabled, a new problem emerges: how to decide whether a motion should be personalized or universal? Without supervision, a native separation from gradient would be highly biased because of uncertainties like initialization and data disparity, making some personalized motions inappropriately stored into the universal field but too much universal knowledge omitted.

For this problem, we introduce a Negative Contrast Loss with two principles: 1) the ideal personalized motions are expected to be the minimum additional motions that enable smooth model convergence, thus the fewer the better; 2) subject to the first point, the personalized motions of one identity should be different from that of the others. Therefore, we design the loss by a truncated dot product:
\begin{equation}
    \label{eq:ncloss}
    \setlength{\abovedisplayskip}{6pt}
    \setlength{\belowdisplayskip}{6pt}
    \mathcal{L}_C(\Delta \mu_P^i, \Delta \mu_P^j) = \mathbb{I}_{trunc} (\Delta \mu_P^i \cdot  \Delta \mu_P^j),
\end{equation}
where $(\Delta \mu_P^i, \Delta \mu_P^j)$ is a pair of personalized $\mu$-deformation from $i$-th and $j$-th personalized fields, queried by the same $\mu_P$ and $\mathbf{C}$ in Eq. (\ref{eq:pdeform}), and $\mathbb{I}_{trunc}$ is a truncation function:
\begin{equation}
    \setlength{\abovedisplayskip}{5pt}
    \setlength{\belowdisplayskip}{5pt}
    \mathbb{I}_{trunc}(x) = \left\{
                                \begin{array}{lr}
                                    x, & x > 0 \\
                                    0, & x\leq0 \\
                                \end{array}
                                \right.
                                .
\end{equation}
This negative-sample-based contrastive loss works from two aspects: 1) when the two participants are close in the 3D space, the loss would encourage them to be either different in direction or less in effect; 2) if the participants are already significantly different, or if one of the participants is close to $\mathbf{0}$, the loss stops pushing them too far.

By contrasting the negative personalized $\Delta \mu$ pairs, the proposed loss can encourage learned personalized motion to be diverse between the different personalized fields, therefore maximizing the filtering to get universal motions.

\vspace{-1mm}
\subsection{Motion-Aligned Adaptation \label{sec:adaptation}}
\vspace{-1mm}
Based on the pre-trained UMF from Sec. \ref{sec:pretrain}, we introduce a \textit{Motion-Aligned Adaptation} strategy to allow fast and high-quality learning of new-identity talking heads from a short video. Different from existing works \cite{shen2022dfrf, ye2024mimictalk, li2023hide} that take motions to guide image generation, we directly drive Gaussians by deformation in the final rendering for more realistic visual quality, bringing a challenge in precisely aligning the pre-trained motion to the new identity's head.

\noindent\textbf{Motion Aligner. } Due to the diversity of human faces, the pre-trained UMF can not natively match all unseen identities suitably. Except for the personalization that we aim to learn, this motion misalignment exists in two aspects: 1) the target facial structure may be biased with the implicit one from pre-training in UMF; 2) the motion of the target identity may have a scale gap from the universal motion.

To this end, we propose to wrap a Motion Aligner on the motion field to learn the adjustment. Specifically, a multi-resolution positional encoder $\mathcal{H}_A$ is first added to store the spatial information. Then, before querying the motion field by $\mu$ from the structure field, a coordinate offset $\Delta \mu_A$ is predicted as bias adjustment:
\begin{equation}
    \label{eq:alignbias}
    \setlength{\abovedisplayskip}{5pt}
    \setlength{\belowdisplayskip}{6pt}
    \Delta \mu_A = \text{MLP}(\mathcal{H}_A(\mu)) \in \mathbb{R}^3.
\end{equation}
After getting the universal deformation $\delta_U$, the $\Delta \mu$ part is multiplied by a scaling factor $\tau_A \in \mathbb{R}^3$ predicted in the same way as Eq. (\ref{eq:alignbias}) for scale alignment, and finally output the adjusted deformation $\delta'_U = \{\Delta \mu', \Delta s, \Delta q\}$, where
\begin{equation}
    \setlength{\abovedisplayskip}{7pt}
    \setlength{\belowdisplayskip}{7pt}
    \Delta \mu' = \Delta \mu \times \tau_A , \quad \Delta \mu \in \mathcal{U}(\mu + \Delta \mu_A, \mathbf{C}).
\end{equation}
Here $\delta'_U$ is used as the final deformation for rendering.

\begin{figure}[t]
    \centering
    \setlength{\abovecaptionskip}{4pt}
    \includegraphics[width=1\linewidth]{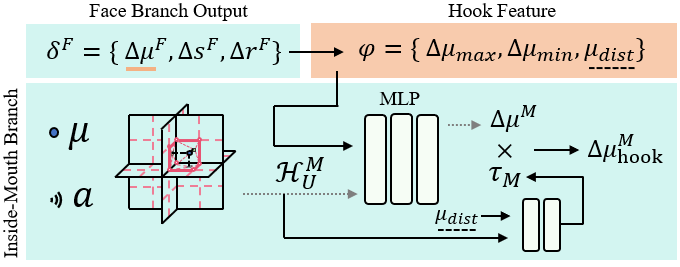}
    \caption{\textbf{Illustration of Face-Mouth Hook}. We hook the motion of the mouth to the generated face motion, allowing an alignment across two branches to enhance robustness under few training data.}
    \label{fig:hook}
    \vspace{-4mm}
\end{figure}

\noindent\textbf{Face-Mouth Hook. }
As a deformation-based framework, the asynchronous inside and out mouth motions often deteriorate the modeling \cite{li2024talkinggaussian}. 
As in Sec. \ref{sec:synthesizer}, we adopt a face-mouth decomposition \cite{li2024talkinggaussian} to solve this problem. It separates the modeling of the face (including lips) and inside of the mouth into two branches.
However, when data is reduced, these two parts may fail to learn to move in harmony, causing misalignment with low visual quality. Without an existing mechanism to ensure the inter-alignment, we suggest setting a hook in the motion field to allow the mouth to actively align to the face, as shown in Figure \ref{fig:hook}.


Given a predicted deformation $\delta_F$ from the face branch that represents face motion, we attempt to fetch the motions on lips from it to guide the inside-mouth deformation $\delta_M$. Considering the lip motion tends to cause the largest deformation amount of the whole face, we take the maximal and minimal $\Delta \mu^F \in \delta_F$ (denoted as $\Delta \mu_{max/min}$) as the motion cues, and calculate a distance $\mu_{dist} = \Delta \mu_{max} - \Delta \mu_{min}$ for a direct scale cue. Then, this set of features $\varphi=\{\Delta \mu_{max}, \Delta \mu_{min}, \mu_{dist}\}$ are input in the MLP decoder in the inside-mouth branch motion field to predict a deformation $\Delta \mu$. The mouth motion field can be formulated as:
\begin{equation}
    \setlength{\abovedisplayskip}{6pt}
    \setlength{\belowdisplayskip}{6pt}
    \Delta \mu^M = \mathcal{U^M}(\mu, \{\boldsymbol{a}, \varphi\}).
\end{equation}
Representing a structure difference, $\mathcal{U^M}$ does not take $\boldsymbol{e}$ as input and does not predict deformations on $r$ and $s$.

Moreover, considering the opening degree of the inside mouth should align with that of the lips, we hook the motion scaling prediction to the facial motion scale cue $\mu_{dist}$. To achieve this, a scaling factor $\tau_M$ is predicted for each $\mu$ by an MLP to get the final primitive-wise deformation:
\begin{equation}
    \setlength{\abovedisplayskip}{8pt}
    \setlength{\belowdisplayskip}{8pt}
    \mu_\text{hook}^M = \Delta \mu^M \times \tau_M, \quad \tau_M = \text{MLP}(\mathcal{H}_U^M(\mu) \oplus \mu_{dist}),
\end{equation}
where $\mathcal{H}_U^M$ is the $\mathcal{H}$ of the UMF in the inside-mouth branch. 

This hook technique establishes a connection between the motion fields to guide the mouth motion to align to the face, while the former's scale is also explicitly connected to the latter, bringing robust performance on the face-mouth alignment even when with only few training data.

\noindent\textbf{Geometry Prior Regularizer. }
To further reduce the geometry ambiguity caused by the lack of view cover, we leverage the power of advanced human geometry estimator \cite{khirodkar2025sapiens} to provide extra geometry cues for regularization. To solve the geometry degradation \cite{li2024dngaussian} and instability of 3DGS at unseen views \cite{cheng2024gaussianpro}, we add two regularizers on the depth and surface normal. Following \cite{cheng2024gaussianpro} we first calculate the 2D depth $D$ and normal $Ns$ when synthesizing image $I$. Then, the estimated monocular depth map $\check{D}$ and normal $\check{N}$ from the ground-truth image are used for regularization:
\vspace{-2mm}
\begin{equation}
    \setlength{\abovedisplayskip}{1pt}
    \setlength{\belowdisplayskip}{5pt}
        \mathcal{L}_{Geo} = \lambda_D L_{D}(D, \check{D}) + \lambda_N \sum_{i=0}^{m}\sum_{j=0}^{n}(1 - N_{i,j} \cdot \check{N}_{i,j}),
\end{equation}
where $L_{D}$ is a scale-invarant depth loss, $N_{i,j}$ denotes the normal at pixel $(i,j)$, and $(m,n)$ is the shape of image $I$.


\begin{figure*}
    \centering
    \setlength{\abovecaptionskip}{5pt}
    \includegraphics[width=0.99\linewidth]{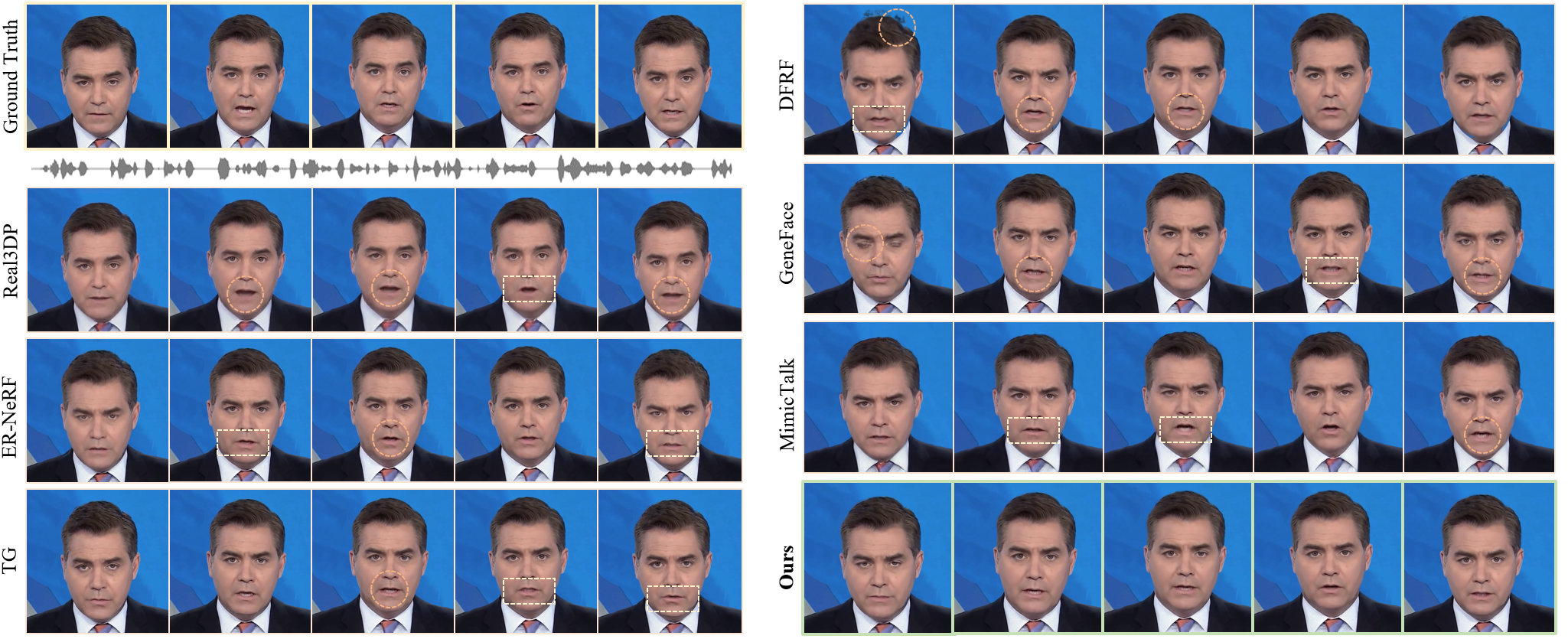}
    \caption{\textbf{Qualitative Comparison on Synchronization.} Our method performs best in both lip-synchronization and visual quality. ``Real3DP" and ``TG" denote \cite{ye2024real3d} and \cite{li2024talkinggaussian}. Better visualized with \textbf{zoom-in}. We recommend watching the \textit{supplementary video}.}
    \label{fig:exp1}
    \vspace{-1mm}
\end{figure*}

\begin{table*}[]
\resizebox{1\linewidth}{!}{
\setlength{\tabcolsep}{2mm}
\centering
\begin{tabular}{@{}l|c|ccc|ccc|cc|c}
\toprule
\,\multirow{2}{*}{Methods} & \multirow{3}{*}{Setting}                                                                             & \multicolumn{3}{c|}{Rendering Quality}                              & \multicolumn{3}{c|}{Motion Quality}                         & \multicolumn{2}{c|}{Efficiency}        & \multirow{3}{*}{Real-time}      \\
                         &                                                                                                      & PSNR (A/F) $\uparrow$  & LPIPS (A/F) $\downarrow$ & SSIM $\uparrow$ & LMD $\downarrow$ & AUE-(L/U) $\downarrow$ & Sync-C $\uparrow$ & Training $\downarrow$ & FPS $\uparrow$ &                                 \\
\,Ground Truth             &                                                                                                      & N/A                    & 0 / 0                    & 1.000           & 0                & 0 / 0                  & 8.897           & -                     & -              &                                 \\ \midrule
\,EAT \cite{gan2023eat}       & \multirow{2}{*}{One-shot}                                                                            & \quad - \quad / 22.44  & \quad - \quad / 0.060    & 0.712           & -                & 1.309 / 1.011          & \textbf{7.577}  & -                     & 11.8           & \textcolor{BrickRed}{\ding{55}} \\
\,Real3DPortrait \cite{ye2024real3d} &                                                                                                      & 23.58* / 25.37          & 0.104* / 0.054            & 0.821           & {\ul 3.407}      & 1.148 / 1.098          & {\ul 7.549}     & -                     & 8.4            & \textcolor{BrickRed}{\ding{55}} \\ \midrule
\,RAD-NeRF \cite{tang2022rad} & \multirow{4}{*}{\begin{tabular}[c]{@{}c@{}}Trained\\ From Scratch\end{tabular}}                        & 28.36 / 25.51          & 0.048 / 0.035            & 0.835           & 3.555            & 1.295 / 0.732          & 2.249           & 5 hours               & 28.6           & \ding{51}                       \\
\,ER-NeRF \cite{li2023efficient}  &                                                                                                      & 28.23 / 25.63          & {\ul 0.040} / 0.031        & 0.844           & 3.541            & 1.327 / 0.451          & 3.074           & 2 hours               & 30.8           & \ding{51}                       \\
\,GaussianTalker \cite{cho2024gaussiantalker} &                                                                                                      & 28.18 / 25.61          & 0.043 / 0.032            & 0.836           & 3.647            & 1.379 / 0.415          & 1.970           & 67 min               & 64.5           & \ding{51}                       \\
\,TalkingGaussian \cite{li2024talkinggaussian} &                                                                                                      & 28.32 / 26.01          & 0.041 / {\ul 0.028}            & 0.843           & 3.588            & 1.158 / {\ul 0.316}     & 3.556           & 31 min                & \textbf{114.2} & \ding{51}                       \\ \midrule
\,DFRF  \cite{shen2022dfrf} & \multirow{4}{*}{\begin{tabular}[c]{@{}c@{}}Adaptation \\ \small \& Few-shot\end{tabular}} & {\ul 28.48} / {\ul 26.26}    & 0.066 / 0.034            & {\ul 0.859}     & 3.436            & 1.247 / 0.683          & 3.957           & 9 hours               & 0.03           & \textcolor{BrickRed}{\ding{55}} \\
\,GeneFace \cite{ye2023geneface} &                                                                                                      & 27.23 / 25.08          & 0.052 / 0.036            & 0.823           & 3.510            & 1.310 / 0.684          & 3.651           & 7 hours               & 18.1           & \textcolor{BrickRed}{\ding{55}}                       \\
\,MimicTalk \cite{ye2024mimictalk} &                                                                                                      & 24.69* / 26.27         & 0.075* / 0.031           & 0.853           & 3.489            & {\ul 0.958} / 0.775  & 6.926           & {\ul 16 min}          & 8.2            & \textcolor{BrickRed}{\ding{55}} \\
\,\textbf{InsTaG (Ours)}   &                                                                                                      & \textbf{28.86} / \textbf{26.32} & \textbf{0.039} / \textbf{0.026}   & \textbf{0.861}  & \textbf{3.167}   & \textbf{0.926 / 0.313} & 5.318           & \textbf{13 min}       & {\ul 82.5}     & \ding{51}                       \\ \bottomrule
\end{tabular}
}
\vspace{-1mm}
\caption{\textbf{Quantitative Comparison in \textit{self-reconstruction setting}} with 5s training data. InsTaG achieves the best image quality and personalized dynamics, with fast adaption and real-time inference. We mark the \textbf{best} and {\ul second-best} results. (*: with synthesized torso.)}
\vspace{-3mm}
\label{tab:setting1_1}
\end{table*}

\subsection{Training Details}
\noindent\textbf{Photometric Loss.}
The training processes are mainly supervised by the image photometric loss $\mathcal{L}_I$ between the synthesized image $I$ and its ground-truth $I_{GT}$. Following 3DGS \cite{kerbl2023gaussian}, $\mathcal{L}_I$ consists of an L1 and a D-SSIM \cite{wang2004ssim} terms.

\noindent\textbf{Pre-training.}
During pre-training, we collect $k$ videos as training data to simultaneously train the UMF. For each synthesized image $I^i$ from $i$-th video, we first calculate its photometric loss, and then calculate the Negative Contrast Loss to each other. The pre-training loss for the $i$-th video is:
\begin{equation}
    \setlength{\abovedisplayskip}{1pt}
    \setlength{\belowdisplayskip}{2pt}
    \mathcal{L}_{pre}^i = \mathcal{L}_I(I^i, I^i_{GT}) + \lambda_{C}\sum_{j=1,j \neq i}^{k}\mathcal{L}_C(\Delta \mu_P^i \cdot  \Delta \mu_P^j).
\end{equation}

\noindent\textbf{Adaptation.}
For adaptation, we initialize a person-specific model with the pre-trained UMF. A warm-up stage first runs without a Motion Aligner and geometry loss for a start. After that, the model is trained with full loss:
\begin{equation}
    \setlength{\abovedisplayskip}{5pt}
    \setlength{\belowdisplayskip}{0pt}
    \mathcal{L}_{ada} = \mathcal{L}_I(I, I_{GT}) + \mathcal{L}_{Geo}(D, N).
\end{equation}

\vspace{-2mm}
\section{Experiments}
\vspace{-1mm}

\noindent\textbf{Dataset.} 
In the experiments, we collect a long-video corpus with 5 long speaking videos from \cite{li2023efficient, ye2023geneface} for pre-training, and follow prevailing settings \cite{guo2021ad, li2023efficient, li2024talkinggaussian, ye2023geneface} to take 4 videos from publicly-released sets \cite{shen2022dfrf, ye2023geneface} for the test, with no identity overlap of the two parts. The videos have an average length of about 800-8000 frames in 25 FPS with a center portrait, all are cropped and resized to $512\times512$. 

\noindent\textbf{Implementation Details.}
Our method is implemented on PyTorch. The pre-training stage lasts $150,000$ iterations and adaptation lasts $10,000$. An AdamW \cite{loshchilov2018adamw} optimizer is used with learning rates of 5e-3 and 5e-4 separately for the grid and neural networks to train UMF. $\lambda_D$, $\lambda_N$ and $\lambda_C$ are set to 1e-2, 1e-3 and 1. All experiments are performed on RTX 3080 Ti GPUs. The overall pre-training takes about 5 hours. The head pose is estimated through BFM \cite{paysan2009bfm}. A DeepSpeech model \cite{hannun2014deepspeech} is used to extract basic audio features following previous works \cite{guo2021ad, li2023efficient, li2024talkinggaussian}.  

\noindent\textbf{Comparison Baselines.}
To compare with the state-of-the-art methods, we take the baselines from the relevant person-specific methods DFRF \cite{shen2022dfrf}, RAD-NeRF \cite{tang2022rad}, GeneFace \cite{ye2023geneface}, ER-NeRF \cite{li2023efficient}, TalkingGaussian \cite{yu2022talkingdiff}, GaussianTalker \cite{cho2024gaussiantalker} and a concurrent method MimicTalk \cite{ye2024mimictalk}. We also involve the one-shot EAT \cite{gan2023eat} and Real3DPortrait \cite{ye2024real3d}. All the methods are reproduced by their official codebase. If officially supported, we use DeepSpeech as audio extactor.

\begin{figure*}[!t]
    \centering
    \setlength{\abovecaptionskip}{3pt}
    \includegraphics[width=0.98\linewidth]{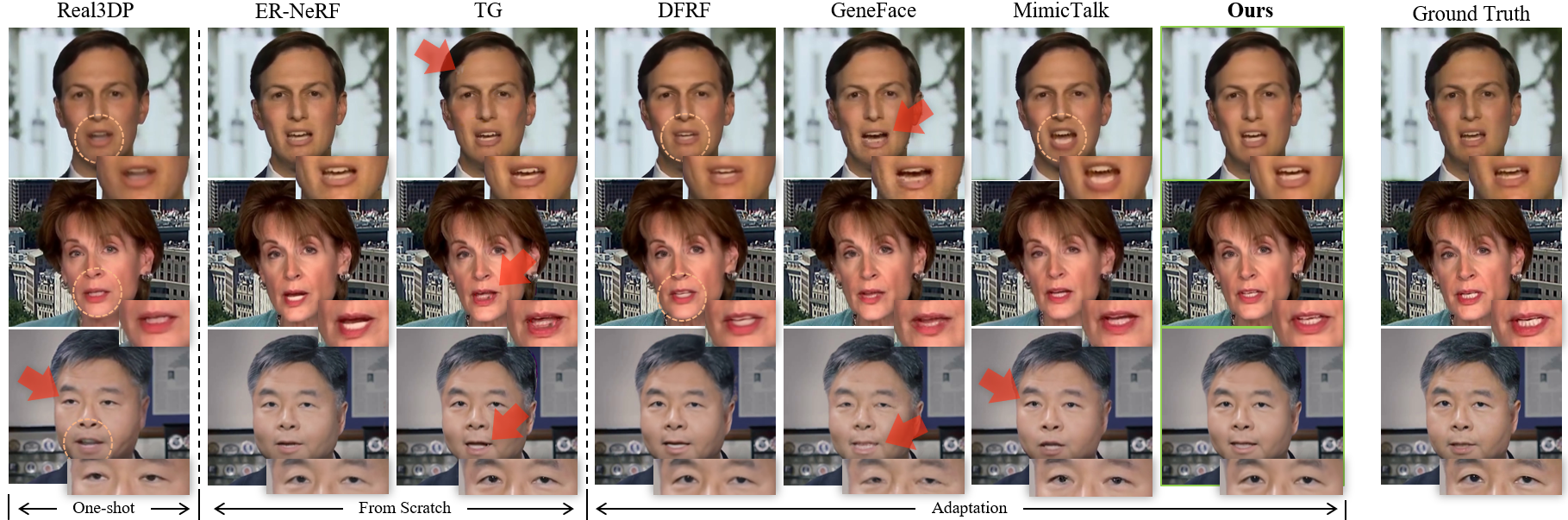}
    \caption{\textbf{Qualitative Comparison on Reconstruction Quality.} Our method performs the best in rendering photorealistic talking heads with fine details. ``Real3DP" and ``TG" denote Real3DPortrait \cite{ye2024real3d} and TalkingGaussian \cite{li2024talkinggaussian}. Better visualized with \textbf{zoom-in}.}
    \label{fig:exp2}
    \vspace{-2mm}
\end{figure*}

\begin{table*}[!t]
\resizebox{1\linewidth}{!}{
\setlength{\tabcolsep}{3.6 mm}
\centering
\begin{tabular}{@{}l|c|ccc|ccc|ccc}
\toprule
\,\multirow{2}{*}{Method} & \multirow{2}{*}{Type}         & \multicolumn{3}{c|}{5s}                                & \multicolumn{3}{c|}{10s}                               & \multicolumn{3}{c}{20s}                                \\ 
                        &                               & PSNR $\uparrow$ & AUE-L $\downarrow$ & Sync $\uparrow$ & PSNR $\uparrow$ & AUE-L $\downarrow$ & Sync $\uparrow$ & PSNR $\uparrow$ & AUE-L $\downarrow$ & Sync $\uparrow$ \\ \midrule
\,ER-NeRF \cite{li2023efficient}     & \multirow{2}{*}{From Scratch} & 28.235          & 1.327              & 3.074           & 28.728          & 1.078              & 3.240           & 29.287          & {\ul 0.941}        & 4.638           \\
\,TalkingGaussian \cite{li2024talkinggaussian} &                               & 28.321          & {\ul 1.158}        & 3.556           & {\ul 29.130}    & 1.069              & 4.663           & {\ul 29.536}    & 1.073              & {\ul 4.776}     \\ \midrule
\,DFRF  \cite{shen2022dfrf}   & \multirow{3}{*}{Adaptation}   & {\ul 28.481}    & 1.247              & {\ul 3.957}     & 28.881          & 1.072          & 4.210           & 29.253          & 1.137              & 4.333           \\
\,GeneFace  \cite{ye2023geneface}  &                               & 28.227          & 1.310              & 3.651           & 27.997          & {\ul 1.036}              & {\ul 4.869}     & 28.825          & 1.033              & 4.772           \\
\,\textbf{InsTaG (Ours)}  &                               & \textbf{28.863} & \textbf{0.926}     & \textbf{5.318}  & \textbf{29.421} & \textbf{0.881}     & \textbf{5.636}  & \textbf{30.352} & \textbf{0.710}     & \textbf{5.764}  \\ \bottomrule
\end{tabular}
}
\vspace{-2mm}
\caption{\textbf{Quantitative Comparison in \textit{self-reconstruction setting}} to person-specific baselines with different data amounts. Our method performs best in various data scenarios, exhibiting robustness. The best and second-best results are in \textbf{bold} and {\ul underline}.}
\vspace{-3mm}
\label{tab:setting1_2}
\end{table*}

\begin{table}[t]
\resizebox{1\linewidth}{!}{
\setlength{\tabcolsep}{2.5 mm}
\centering
\begin{tabular}{@{}lcccc@{}}
\toprule
\,\multirow{2}{*}{Methods} & \multicolumn{2}{c}{Audio \small \textit{English, Male}}       & \multicolumn{2}{c}{Audio \small \textit{German, Female}}        \\ \cmidrule(l){2-3} \cmidrule(l){4-5} 
                         & Sync-E $\downarrow$ & Sync-C $\uparrow$ & Sync-E $\downarrow$ & Sync-C $\uparrow$ \\
\,Full Baseline \cite{li2024talkinggaussian}     & 10.443              & 4.342             & 9.942               & 4.949             \\ \midrule
\,ER-NeRF \cite{li2023efficient}                        & 11.711              & 3.043             & 11.502              & 2.853             \\
\,TalkingGaussian \cite{li2024talkinggaussian}   & 10.675              & 4.317             & 10.898              & 3.708             \\
\,DFRF \cite{shen2022dfrf}                       & 10.515              & 4.726             & {\ul 10.482}        & {\ul 4.329}       \\
\,GeneFace \cite{ye2023geneface}                 & {\ul 9.995}         & {\ul 4.734}       & 11.029              & 4.162             \\ \midrule
\,\textbf{InsTaG (Ours)}   & \textbf{9.886}      & \textbf{4.828}    & \textbf{9.733}      & \textbf{4.990}    \\ \bottomrule
\end{tabular}
}
\vspace{-2mm}
\caption{\textbf{Comparison in \textit{cross-domain setting}} with 20s training data. We take TalkingGaussian \cite{li2024talkinggaussian} with full data for reference.}
\vspace{-5mm}
\label{tab:setting2_1}
\end{table}

\begin{table*}[t]
\resizebox{1\linewidth}{!}{
\setlength{\tabcolsep}{2.2mm}
\centering
\begin{tabular}{@{}lccccccc@{}}
\toprule
\,Methods             & ER-NeRF \cite{li2023efficient} & \footnotesize TalkingGausian \cite{li2024talkinggaussian} & \footnotesize \,GaussianTalker \cite{cho2024gaussiantalker} & DFRF \cite{shen2022dfrf} & GeneFace \cite{ye2023geneface} & MimicTalk \cite{ye2024mimictalk} & \textbf{InsTaG (Ours)} \\ \midrule
\,Lip-sync Accuracy   & 2.00     & 1.90            & 1.50            & 3.30  & 2.70      & \textbf{4.80} & {\ul 4.50}     \\
\,Image Quality       & 3.40     & 3.70            & {\ul 4.00}      & 2.90  & 3.10      & 3.70          & \textbf{4.80}  \\
\,Identity Preserving & 2.40     & 1.50            & 1.70            & 2.90  & 3.00      & {\ul 4.20}    & \textbf{4.60}  \\
\,Video Realness      & 1.40     & 1.50            & 1.40            & 2.80  & 1.70      & {\ul 3.20}    & \textbf{3.90}  \\ \bottomrule
\end{tabular}
}
\vspace{-2mm}
\caption{\textbf{User Study.} The rating is in the range of 1-5, higher denotes better. The best and second-best results are in \textbf{bold} and {\ul underline}.}
\vspace{-4mm}
\label{tab:userstudy}
\end{table*}

\subsection{Evaluation}

\noindent\textbf{Comparison Settings.}
The quantitative comparison contains two settings: \textbf{1)} The \emph{self-reconstruction setting}. In this setting, we split each of the 4 adaptation videos into several few-second training sets and a test set of at least 12s, and use the unseen test data as input and ground truth for evaluation.  \textbf{2)} The \emph{cross-domain setting}, where we use an unseen English male audio from NVP \cite{thies2020nvp} and a German female one from \cite{guo2021ad} to test the cross-domain generalizability. To fully evaluate the baselines with failure models, we use 20s from the only English female video to train one model per method that represents the best performance for testing.

\noindent\textbf{Metrics and Measurements.}
We employ PSNR, LPIPS \cite{zhang2018lpips}, and SSIM \cite{wang2004ssim} to evaluate image quality. Especially, we report PSNR and LPIPS separately on an all image (``A") and only the aligned face (``F"). SSIM is only measured on the face.  For lip synchronization, we use LMD \cite{chen2018lmd} and SyncNet \cite{chung2017syncnet1, chung2017syncnet2} confidence (Sync-C) and error (Sync-E). Besides, action unit \cite{Ekman1978FacialAC} error (AUE) is used on upper-face (AUE-U) and lower-face (AUE-L) following \cite{li2024talkinggaussian}, in which AUE-L can reflect the similarity of the personalized talking actions from the target. We also statistic the training time (fine-tuning time for few-shot or adaptation-based methods) and inference FPS to evaluate efficiency.

\noindent\textbf{Quantitative Evaluation.}
\textbf{1)} Table \ref{tab:setting1_1} reports the results of \emph{self-reconstruction setting} with 5s training data. Our method achieves the best image quality, demonstrating our ability to faithfully generate a realistic talking head. Inheriting the traditions of large corpus pre-training, the one-shot methods \cite{gan2023eat, ye2024real3d} and MimicTalk \cite{ye2024mimictalk} achieve very high Sync scores but are relatively slow. Also getting a high Sync, InsTaG achieves the best in LMD and AUE-L, showing our advantage in preserving personalization. While all baselines can not simultaneously achieve high synchronization and real-time inference, our method produces the best overall performance including the fastest adaptation and high speed. 
\textbf{2)} When data increases, the results in Table \ref{tab:setting1_2} show that InsTaG still outperforms other person-specific models in all aspects, demonstrating our wider application value and robustness to various scenarios.
\textbf{3)} In the results from Table \ref{tab:setting2_1} in \emph{cross-domain setting}, where the methods are tested on two difficult cross-domain scenarios, the advantages of pre-training are further amplified. GeneFace \cite{ye2023geneface} and DFRF \cite{shen2022dfrf} surpass all models trained from scratch. At the same time, InsTaG exceeds the baseline \cite{li2024talkinggaussian} trained on about 1-minute full training data, demonstrating the outstanding robustness and generalizability of our method.

\noindent\textbf{Qualitative Evaluation.}
To qualitatively evaluate the results, we show the generated sequences and rendering details from \textit{self-reconstruction setting}. In Figure \ref{fig:exp1}, it can be observed that the models trained from scratch (ER-NeRF, TG) \cite{li2024er, li2024talkinggaussian} behave poorly in synchronization (yellow box), while the adaptation-based methods (DFRF, GeneFace, MimicTalk) \cite{shen2022dfrf, ye2023geneface, ye2024mimictalk} perform better but generate blurry results (orange circle). Referring to Figure \ref{fig:exp2}, while all methods synthesize either blurs or heavily distorted details (red arrow), our method can reconstruct photorealistic appearance and delicate details. By benefitting from the advantages of both pre-training and person-specific training, InsTaG achieves the best in both synchronization and image quality. We strongly recommend watching our \textit{supplementary video} for an intuitive comparison.

\noindent\textbf{User Study.} 
To better judge the quality in real human-participated scenarios, we conducted a user study with 28 generated talking head videos from 7 methods with 5s training data. We invited 10 attendees to rate each anonymous method from four aspects: (1) Lip-sync Accuracy; (2) Video Realness; (3) Identity Preserving and (4) Video Realness. The results in Table \ref{tab:userstudy} show that our InsTaG achieves the best in three aspects and the second-best in lip-sync, which is only after the concurrent MimicTalk \cite{ye2024mimictalk}, demonstrating our outstanding visual quality via human judgment.

\begin{table}[t]
\resizebox{1\linewidth}{!}{
\setlength{\tabcolsep}{1.5 mm}
\centering
\begin{tabular}{@{}cc|cc|cccc@{}}
\toprule
\multicolumn{2}{c|}{Pre-training}  & \multicolumn{2}{c|}{Adaptation} & \multirow{2}{*}{PSNR $\uparrow$} & \multirow{2}{*}{LPIPS $\downarrow$} & \multirow{2}{*}{LMD $\downarrow$} & \multirow{2}{*}{Sync-C $\uparrow$} \\
PField     & NCLoss                & Aligner        & Hook           &                                  &                                     &                                   &                                    \\ \midrule
           &                       &                &                & 28.66                            & 0.044              
                 & 3.734                             & 2.019                              \\ \midrule
\checkmark &                       &                &                & 28.76                            & 0.040                               & 3.239                             & 4.816                              \\
\checkmark & \checkmark            &                &                & 28.59                            & 0.041                               & {\ul 3.223}                       & 5.124                              \\ \midrule
\checkmark & \checkmark            & \checkmark     &                & 28.69                            & {\ul 0.040}                         & 3.274                             & {\ul 5.233}                        \\
\checkmark & \checkmark            &                & \checkmark     & 28.66                            & 0.040                               & 3.257                             & 5.086                              \\
\checkmark &                       & \checkmark     & \checkmark     & {\ul 28.79}                      & 0.040                               & 3.226                             & 4.994                              \\  \midrule
\checkmark & \checkmark            & \checkmark     & \checkmark     & \textbf{28.86}                   & \textbf{0.039}                      & \textbf{3.167}                    & \textbf{5.318}                     \\ \bottomrule
\end{tabular}
}
\vspace{-2mm}
\caption{\textbf{Ablation Study.} We verify the components under 5s \textit{self-reconstruction} setting, showing the effect of our contributions.}
\label{tab:ablation}
\vspace{-5mm}
\end{table}

\subsection{Ablation Study}

We conduct the ablation study to prove the effectiveness of our contributions. The results are reported in Table \ref{tab:ablation}.

\vspace{0pt}\noindent\textbf{Identity-Free Pre-training.} 
In the first row, we show the results pre-trained without Personalized Fields (PField), which produces bad synchronization due to the identity-conflict problem. After applied PField, the pre-training works with obvious effect. Combined with PField, Negative Contrast Loss (NCLoss) shows its impact on lip-sync improvement, offering higher Sync-C. This demonstrates the necessity of PFiled in enabling the pre-training and the effect of NCLoss on encouraging knowledge collection.

\begin{table}[t]
\resizebox{1\linewidth}{!}{
\setlength{\tabcolsep}{3.2 mm}
\centering
\begin{tabular}{@{}l|c|cccc@{}}
\toprule
\,Backbone           & Hook       & PSNR $\uparrow$ & LPIPS $\downarrow$ & LMD $\downarrow$ & Sync-C $\uparrow$ \\ \midrule
\,Single Branch      & -          & 28.59           & 0.049              & {\ul 3.233}      & 4.675             \\
\,Face-Mouth Decomp. & $\times$   & {\ul 28.69}     & {\ul 0.040}        & 3.274            & {\ul 5.233}       \\
\,Face-Mouth Decomp. & \checkmark & \textbf{28.86}  & \textbf{0.039}     & \textbf{3.167}   & \textbf{5.318}    \\ \bottomrule
\end{tabular}
}
\vspace{-2mm}
\caption{\textbf{Ablation Study on Mouth Motion Representation.} Introducing FMD \cite{li2024talkinggaussian} can effectively improve lip-synchronization, while our FM hook technique further boosts image quality.}
\label{tab:ablation_2}
\vspace{-2mm}
\end{table}

\begin{figure}[t]
    \centering
    \setlength{\abovecaptionskip}{3pt}
    \includegraphics[width=0.95\linewidth]{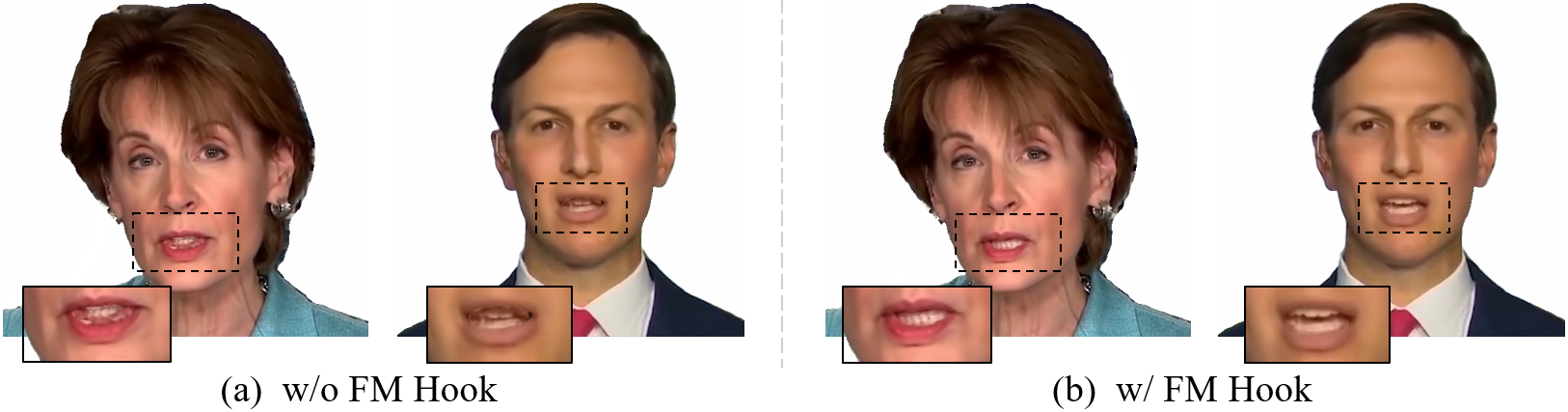}
    \caption{\textbf{Visualization Results of Face-Mouth Hook.} FMD improves performance but also causes misalignment when with inadequate data, which can be well solved by our Face-Mouth Hook.}
    \label{fig:ab_hook}
    \vspace{-5mm}
\end{figure}

\vspace{0pt}\noindent\textbf{Motion-Aligned Adaptation.} We apply our Motion Aligner (Aligner) and Face-Mouth Hook (FM Hook) to illustrate their effect on adaptation. Although the performance is improved by applying our pre-training strategy, the Aligner can bring extra enhancement by providing adaptation adjustments. Furthermore, the Hook works in both visual quality and dynamics by correcting inside-mouth misalignment. Combining these two adaptation techniques, the performance achieves the best among all settings.

\vspace{0pt}\noindent\textbf{Mouth Motion Representation.}
To further explain the necessity of Face-Mouth Decomposition (FMD) \cite{li2024talkinggaussian} and our FM Hook, we take studies by using one single branch to represent the two parts. Results in Table \ref{tab:ablation_2} show the effect of FMD. However, simply introducing it would cause significant misalignment that lowers the visual quality, as in Figure \ref{fig:ab_hook}. By applying our FM Hook, this drawback can be well solved to benefit FMD with robust performance, generating synchronized talking heads with high visual quality.


\vspace{-1mm}
\section{Ethical Consideration}
\vspace{-1mm}
We hope our method can promote the healthy development of digital industries. However, it is important to note that it could be misused for malicious purposes, potentially causing negative impacts. As part of our responsibility, we will assist in developing deepfake detection tools. We recommend the responsible use of this and all similar techniques.

\newpage

\maketitlesupplementary

\section*{Overview}
In the supplementary material, we first report additional experiments in Sec. \ref{sec:exp}, and describe the details of the dataset and implementation in Sec. \ref{sec:data} and \ref{sec:impl}. We further show the visualized result in Sec. \ref{sec:vis}, and discuss the ethical consideration and limitations in Sec. \ref{sec:ethi} and \ref{sec:discuss}. For a better explanation, a supplementary video is additionally provided.

\section{Additional Experiments}
\label{sec:exp}

\subsection{Ablation Study}

\begin{figure}[h]
    \centering
    \includegraphics[width=1\linewidth]{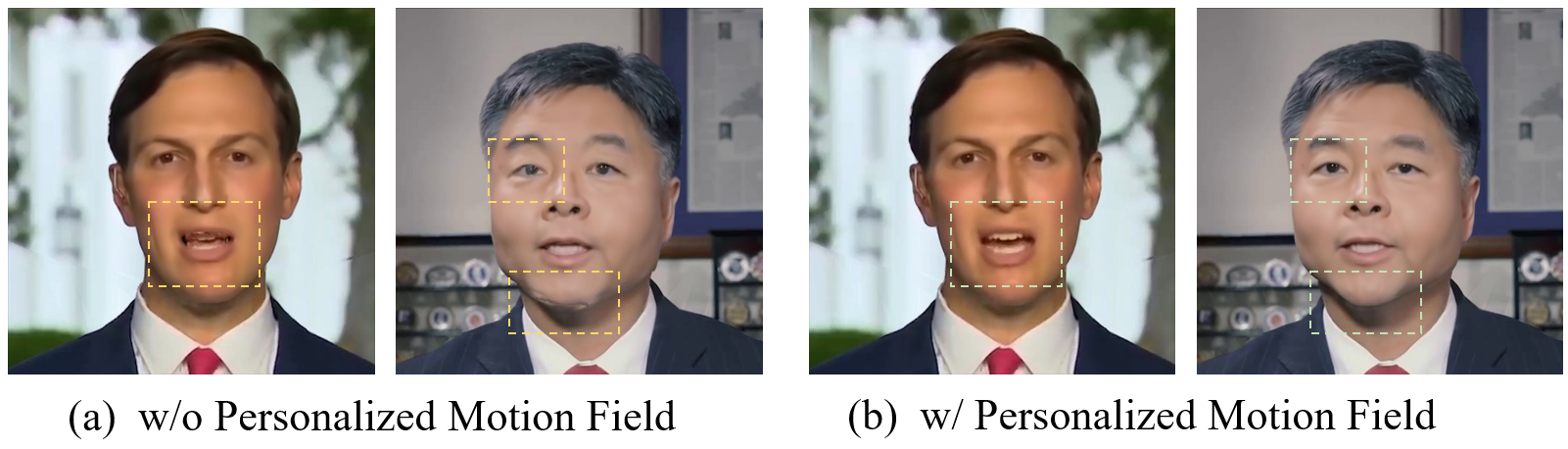}
    \caption{\textbf{Visualized Results of Personalized Motion Field.} The overfitting of personalized motions from multiple persons pollutes the pre-trained module and leads to unexpected distortions on new identities, which can be well solved by adding a motion field in the Personalized Field during pre-training.}
    \label{fig:ablation_pf}
\end{figure}

\paragraph{Personalized Field.}
To further verify the effect of the Personalized Field in pre-training, we conduct a fine-grained ablation study to separately show the effect of the structure field and motion field in it. The results are reported in Table \ref{tab:ablation_pf}. Due to the identity-conflict problem discussed in Sec. 3.2, it can be observed that the pre-training totally fails in helping adaptation when without the personalized structure field, resulting in very low Sync-C scores. Though a significant performance raise happens after applying structure fields, the image quality and synchronization are still not ideal because the heavily overfitted pre-training model is hard to adjust well during the fine-tuning, resulting in unexpected distortions, as shown in Figure \ref{fig:ablation_pf}. After applying the full Personalized Fields, our method achieves the best performance, demonstrating the effect of the two parts.

\begin{table}[h]
\centering
\resizebox{\linewidth}{!}{%
\begin{tabular}{@{}cc|cccc@{}}
\toprule
Structure  & Motion     & PSNR $\uparrow$ & LPIPS $\downarrow$ & LMD $\downarrow$ & Sync-C $\uparrow$ \\ \midrule
\ding{55}  & \ding{55}  & {\ul 28.67}     & 0.044              & 3.734            & 2.019             \\
\ding{55}  & \checkmark & 28.63           & 0.045              & 3.715            & 2.312             \\
\checkmark & \ding{55}  & 28.59           & {\ul 0.040}        & {\ul 3.228}      & {\ul 4.975}       \\
\checkmark & \checkmark & \textbf{28.86}  & \textbf{0.039}     & \textbf{3.167}   & \textbf{5.318}    \\ \bottomrule
\end{tabular}%
}
\caption{\textbf{Ablation Study on Personalized Field.} The best and second-best results are highlighted with \textbf{bold} and {\ul underline}.}
\label{tab:ablation_pf}
\vspace{-3mm}
\end{table}

\paragraph{Motion Aligner.}
We take an additional ablation study inside the Motion Aligner to illustrate the effect of the two involved alignments. As shown in Table \ref{tab:ablation_aligner}, when not applying the offset alignment $\Delta \mu_A$ from Eq. (8), the performances are relatively worse on Sync-C with a lower PSNR. This comes from the insufficient inheritance of the pre-trained knowledge due to the head structure misalignment. After adding the offset adjustment, Sync-C and PSNR increase significantly. Combined with the scale adjustment $\tau_A$ from Eq. (9), our method finally achieves an ideal performance.

\begin{table}[h]
\centering
\setlength{\tabcolsep}{3mm}
\resizebox{\linewidth}{!}{%
\begin{tabular}{@{}cc|cccc@{}}
\toprule
Offset     & Scale      & PSNR $\uparrow$ & LPIPS $\downarrow$ & LMD $\downarrow$ & Sync-C $\uparrow$ \\ \midrule
\ding{55}  & \ding{55}  & 28.66           & 0.040              & 3.257            & 5.086             \\
\ding{55}  & \checkmark & 28.68           & 0.040              & 3.211            & 5.113             \\
\checkmark & \ding{55}  & 28.77           & {\ul 0.040}        & {\ul 3.192}      & {\ul 5.229}       \\
\checkmark & \checkmark & \textbf{28.86}  & \textbf{0.039}     & \textbf{3.167}   & \textbf{5.318}    \\ \bottomrule
\end{tabular}%
}
\caption{\textbf{Ablation Study on Motion Aligner.} The best and second-best results are highlighted with \textbf{bold} and {\ul underline}.}
\label{tab:ablation_aligner}
\vspace{-4mm}
\end{table}

\begin{figure}[h]
    \centering
    \includegraphics[width=1\linewidth]{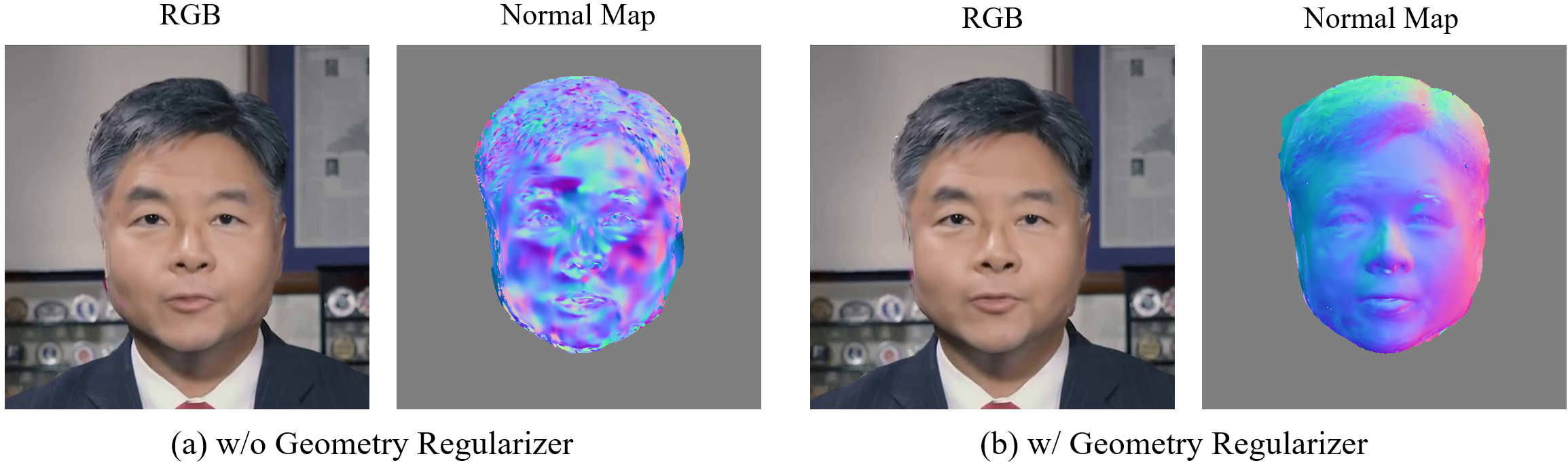}
    \caption{Due to the limited view coverage, the geometry of the 3D talking head degrades, leading to inaccurate surfaces.}
    \label{fig:georeg1}
\end{figure}

\begin{figure}[h]
    \centering
    \setlength{\abovecaptionskip}{6pt}
    \includegraphics[width=1\linewidth]{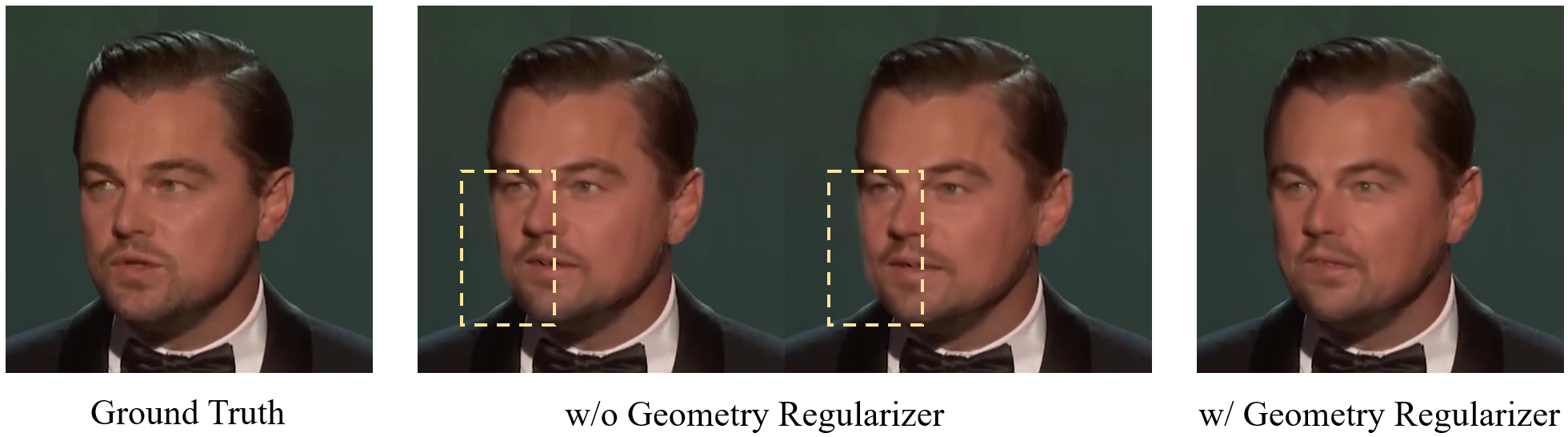}
    \caption{By introducing geometry constraints, our method successfully solves the serious artifacts in difficult unseen views and thus provides a clearer talking head with accurate structure.}
    \label{fig:georeg}
    \vspace{-2mm}
\end{figure}

\paragraph{Geometry Prior Regularizer.}
Scenes purely reconstructed by radiance fields have excellent visual quality at interpolated viewpoints. However, they are weak in extrapolation, which corresponds to the situations where the talking heads rotate to an unseen angle during synthesis. One main reason is that the geometry is degraded due to the limited view coverage, as shown in Figure \ref{fig:georeg1}. To tackle this problem, we use the Geometry Prior Regularizer to provide extra constraints to help the 3D head reconstruction. Since these troubles often appear only in a few frames and influence slightly on the overall numerical metrics, we reveal them quantitatively to verify the effect of the regularizer. As shown in Figure \ref{fig:georeg}, the regularizer successfully tackles the mist-like artifacts due to an inaccurate surface location, and thus brings better visual quality.

\begin{table*}[!t]
\centering
\setlength{\tabcolsep}{5mm}
\resizebox{\linewidth}{!}{%
\begin{tabular}{lccccccc}
\toprule
\multirow{2}{*}{Methods}                & \multirow{3}{*}{Extractor} & \multicolumn{3}{c}{Rendering Quality}                  & \multicolumn{3}{c}{Motion Quality}                          \\ \cmidrule(l){3-5} \cmidrule(l){6-8} 
                                        &                            & PSNR $\uparrow$ & LPIPS $\downarrow$ & SSIM $\uparrow$ & LMD $\downarrow$ & AUE-(L/U) $\downarrow$ & Sync-C $\uparrow$ \\
Ground Truth                            &                            & N/A             & 0                  & 1.000           & 0                & 0 / 0                  & 8.897           \\ \midrule
RAD-NeRF \cite{tang2022rad}                          & Wav2Vec 2.0                & 27.42           & 0.050              & 0.831           & 3.898            & 1.437 / 0.879          & 2.299           \\
ER-NeRF \cite{li2023efficient}                & Wav2Vec 2.0                & 27.87           & 0.041              & 0.848           & 3.624            & 1.427 / 0.610          & 2.856           \\
TalkingGaussian \cite{li2024talkinggaussian}        & Wav2Vec 2.0                & 28.13           & 0.041              & 0.850           & 3.612            & 1.226 / 0.400          & 3.635           \\ \midrule
GeneFace \cite{ye2023geneface}             & HuBERT                     & 28.23           & 0.052              & 0.823           & 3.510            & 1.310 / 0.684          & 3.651           \\
ER-NeRF \cite{li2023efficient}         & HuBERT                     & 27.56           & 0.043              & 0.832           & 3.827            & 1.595 / 0.529          & 1.853           \\
TalkingGaussian \cite{li2024talkinggaussian}    & HuBERT                     & 28.23           & 0.043              & 0.840           & 3.860            & 1.312 / 0.467          & 3.367           \\ \midrule
\multirow{3}{*}{\textbf{InsTaG (Ours)}} & DeepSpeech                 & 28.86           & \textbf{0.039}     & {\ul 0.861}     & {\ul 3.167}      & {\ul 0.926} / \textbf{0.313}    & {\ul 5.318}     \\
                                        & HuBERT                     & \textbf{29.00}  & {\ul 0.040}        & 0.861           & 3.174            & 1.031 / {\ul 0.388}          & 5.239           \\
                                        & Wav2Vec 2.0                & {\ul 28.89}     & 0.041              & \textbf{0.867}  & \textbf{3.111}   & \textbf{0.924} / 0.463 & \textbf{5.852}  \\ \bottomrule
\toprule
TalkingGaussian \cite{li2024talkinggaussian}    & AVE                     & 28.52           & 0.041              & 0.845           & 3.075            & 0.836 / 0.631          & \textbf{8.047}           \\ 
\textbf{InsTaG (Ours)}                          & AVE                     & \textbf{28.83}  & \textbf{0.040}     & \textbf{0.852}  & \textbf{2.984}   & \textbf{0.742 / 0.365} & 7.787           \\ 

\bottomrule

\end{tabular}%
}
\caption{\textbf{Comparison with Different Audio Feature Extractors} using 5s training video. The best and second-best results are highlighted with \textbf{bold} and {\ul underline}. After applying an advanced extractor, our method gets further improvements, showing the scaling potential.}
\label{tab:audextractor}
\end{table*}

\begin{figure}[t]
    \centering
    \setlength{\abovecaptionskip}{6pt}
    \includegraphics[width=1\linewidth]{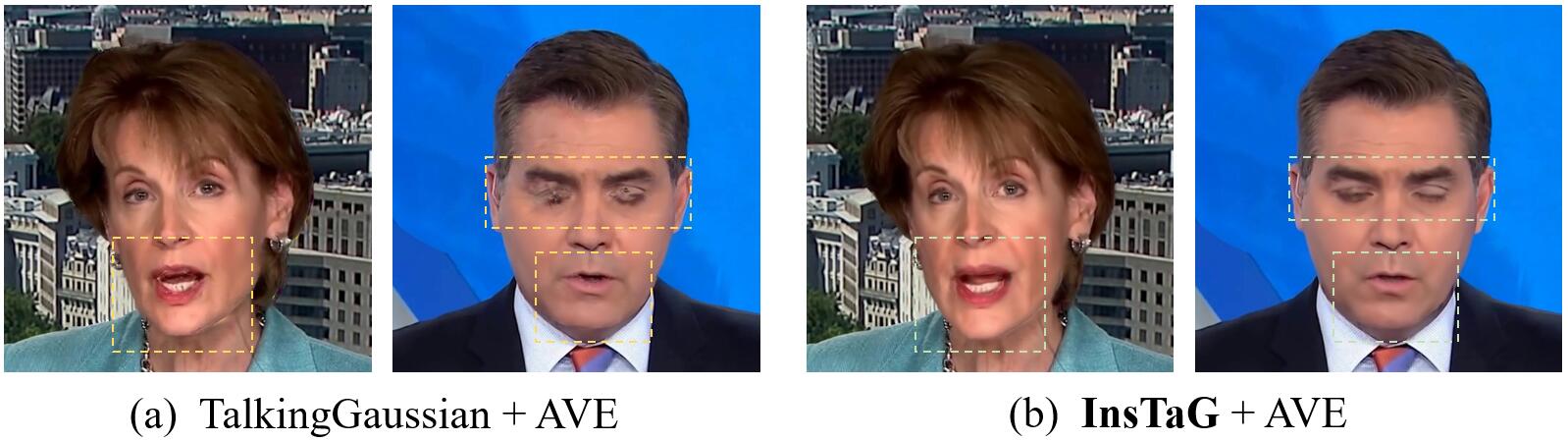}
    \caption{\textbf{Comparision with AVE audio extractor.} Our InsTaG shows advantage in rendering quality and robustness.}
    \label{fig:avecomp}
    \vspace{-3mm}
\end{figure}

\subsection{Audio Feature Extractor}
In the main paper, we use DeepSpeech \cite{hannun2014deepspeech} as the audio feature extractor to align with most previous works  \cite{shen2022dfrf, li2023efficient, li2024talkinggaussian} for a fair comparison. Besides, our InsTaG can also apply other advanced audio extractors like in previous radiance-field-based methods to get a performance improvement. As shown in Table \ref{tab:audextractor}, when applying Wav2Vec 2.0 \cite{baevski2020wav2vec} and HuBERT \cite{hsu2021hubert}, our method gains improvements in different aspects. With the help of Wav2Vect, InsTaG achieves a high Sync-C score and better image quality as well compared to DeepSpeech. HuBERT mainly helps in image quality especially on PSNR, while the lip synchronization drops a bit. Interestingly, we observe this phenomenon also appears on other baselines like ER-NeRF and TalkingGaussian. We consider this because its high feature dimension brings extra difficulty in training with few data. While the baselines still fail to produce reasonable results with a short training video, our method performs best in all aspects.

Additionally, we later noticed that the AVE encoder in SyncTalk \cite{peng2023synctalk} may perform a similar effect as our method, which is pre-trained on large corpus with a lip-sync expert discriminator. Therefore, the extracted signal is much suitable for the task of talking head synthesis. Here we add a comparison between TalkingGaussian and our method in Table \ref{tab:audextractor}. As shown, AVE can exactly boost the results and achieves an extremely high Sync-C, surpassing even the one-shot methods in Table 1. However, TalkingGaussian fails to produce high-fidelity images. Without a comprehensive 3D motion prior, the rendered images often have heavy noises and the motions are jittering, leading to worse rendering, as visualized in Figure \ref{fig:avecomp}. According to the AUE and LMD, it can be observed that the reconstructed talking head is suboptimal with insufficient personalization, while it also can't enable the pre-training of upper-face expressions. While equipped with the stronger AUE, our method can even get a more comprehensive performance, especially in Sync-C that is weak with ASR extractors, surpassing all the baselines in Table 1.

\begin{table}[t]
\resizebox{\columnwidth}{!}{%
\begin{tabular}{lcc|ccccc}
\toprule
Item      & Infer. & Adapt. & P-1  & P-2  & P-3  & P-4  & P-5  \\ \midrule
VRAM (MB) & 1826   & 2214   & 2200 & 2188 & 2194 & 2204 & 2270 \\ \bottomrule
\end{tabular}%
}
\caption{\textbf{Peak GPU Memory Cost.} P-$k$ denotes the pre-training process with $k$ personalized fields involved. The results show the high efficiency of InsTaG and its huge scaling potential.}
\label{tab:effi}
\end{table}

\subsection{Efficiency Analysis}
Besides the inference FPS and training time, we additionally report the GPU memory cost to show the efficiency of our method. From Table \ref{tab:effi}, it can be observed that our InsTaG is efficient in all processes including inference (infer.),  adaptation (adapt.), and pre-training (P-$k$) with multiple personalized fields. Notably, since the personalized fields cost only little storage, the GPU memory cost increases slightly along with the number of fields involved. Thus, there is still a huge space to enhance the performance by adding more pre-training data.

\subsection{More Training Data}
We further evaluate the performance with more training data, which is similar to the setting in previous person-specific methods \cite{guo2021ad, li2023efficient, li2024talkinggaussian, tang2022rad}. As shown in Table \ref{tab:more}, InsTaG still outstands by achieving higher lip synchronization in a shorter time. This shows the common knowledge from pre-training can not only work in few-shot settings but also provide a universal improvement for various data scenarios. 

\begin{table}[t]
\resizebox{\columnwidth}{!}{%
\begin{tabular}{lccccc}
\toprule
Method               & Time            & PSNR $\uparrow$ & LPIPS $\uparrow$ & LMD $\uparrow$ & Sync-C $\uparrow$ \\ \midrule
ER-NeRF              & 2 hours         & 30.07           & {\ul 0.0279}     & 3.111          & 5.088             \\
TalkingGaussian      & 34 min          & 31.04           & \textbf{0.0267}  & 3.001          & 6.005             \\
\textbf{InsTaG-10K}   & \textbf{13 min} & \textbf{31.37}  & 0.0312           & {\ul 2.909}    & {\ul 6.005}       \\
\textbf{InsTaG-25K} & {\ul 32 min}    & {\ul 31.15}     & 0.0305           & \textbf{2.893} & \textbf{6.143}    \\ \bottomrule
\end{tabular}%
}
\caption{\textbf{Comparison with Full Training Data.} We use full of the training data to evaluate the performance. 10K and 25K denote trained with 10,000 and 25,000 iterations.}
\label{tab:more}
\end{table}

\subsection{Comparision with DDPM Baselines}
Here we add two current SOTA DDPM-based methods MEMO \cite{zheng2024memo} and Ditto \cite{li2024ditto}, which support 512$\times$512 generation, for Table 1's comparison. Since they can not reenact the head trace, we report the image quality metrics of the aligned face. While these generative models perform worse in personalization (AUE), visual quality, and speed, InsTaG shows superior overall performance with real-time inference capability. 

\begin{table}[h]
\resizebox{\linewidth}{!}{%
\setlength{\tabcolsep}{2mm} 
\begin{tabular}{lccccccc}
\toprule
Methods       & PSNR $\uparrow$ & LPIPS $\downarrow$ & SSIM $\uparrow$ & LMD $\downarrow$ & AUE-(L/U) $\downarrow$ & Sync-C $\uparrow$ & FPS $\uparrow$ \\ \midrule
MEMO          & 21.21           & 0.056              & 0.742           & -                & 1.20 / 1.10            & \textbf{8.63}     & 0.4           \\
Ditto         & 20.21           & 0.065              & 0.710           & -                & 1.52 / 0.94            & 6.63              & 10.5           \\ \midrule
\textbf{InsTaG} & \textbf{26.32}  & \textbf{0.026}     & \textbf{0.861}  & \textbf{3.167}   & \textbf{0.93 / 0.31}   & 5.32              & \textbf{82.5}  \\ \bottomrule
\end{tabular}%
}
\caption{\textbf{Comparison with DDPM Baselines.} InsTaG shows superior overall performance with real-time inference capability.}
\end{table}

\subsection{3D Effect under Extreme Viewing Angles}
Here, we visualize the learned 3D heads from \textit{5s} data under extreme unseen angles. The results show our learned 3D geometry consistency can be well kept under extreme views, despite some texture degradation due to a lack of data coverage. Notably, the only few training views are mostly frontal, showing our effectiveness and robustness in preserving 3D head structure.
\begin{figure}[h]
    \centering
    \includegraphics[width=1\linewidth]{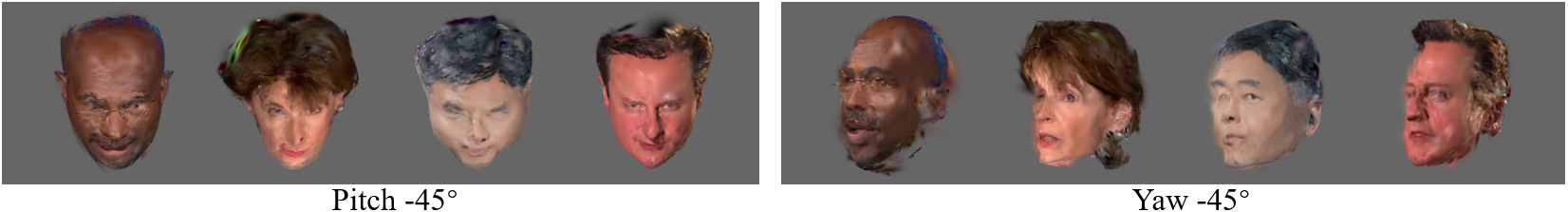}
    \caption{\textbf{3D Effect under Extreme Viewing Angles.} Our learned 3D geometry consistency can be well kept under extreme views using only limited data mostly from the front.}
\end{figure}

\section{Dataset Declaration}
\label{sec:data}
In the main experiments, all the multimedia data were obtained from existing works \cite{shen2022dfrf, li2023efficient, ye2023geneface, thies2020nvp}. These data were collected from publicly released Internet content. The data used in this work contain only public figures to avoid invading personal privacy. All the data are manually checked to reduce the existence of offensive content.

Specifically, we use 5 ready-made celebrity speaking video clips (named "Obama1", "Jae-in", "Shaheen", "may", "macron") from \cite{ye2023geneface, li2024talkinggaussian} as the pre-training data, and take the first 5000 frames for each to pre-train the model. To keep fairness, we do not involve videos with an overlapping celebrity in the above pre-training data in evaluation. With such a principle, following the prevailing setting in previous methods \cite{li2023efficient, li2024talkinggaussian, peng2023synctalk}, we use 4 ready-made videos from \cite{shen2022dfrf, ye2023geneface} as the testing set. To ensure a sound evaluation, we keep at least a clip of 12 seconds for each video for the test in the \textit{self-reconstruction} setting.

In the supplementary video, we also report several models trained with videos collected by ourselves to demonstrate the performance of InsTaG. Since they haven't been used publicly in previous works or lack a provided preprocessing, we do not involve them in the main experiments, therefore to keep the fairness and soundness.

\paragraph{Diversity of races, genders, ages, and facial features.}
We attach great importance to the diversity and try our best to cover different races and genders. However, due to the difficulty in data collection, there are still some lacks. Nevertheless, since our pre-training strategy is identity-free, this negative influence can be weakened.

\begin{figure*}
    \centering
    \vspace{1cm}
    \includegraphics[width=1\linewidth]{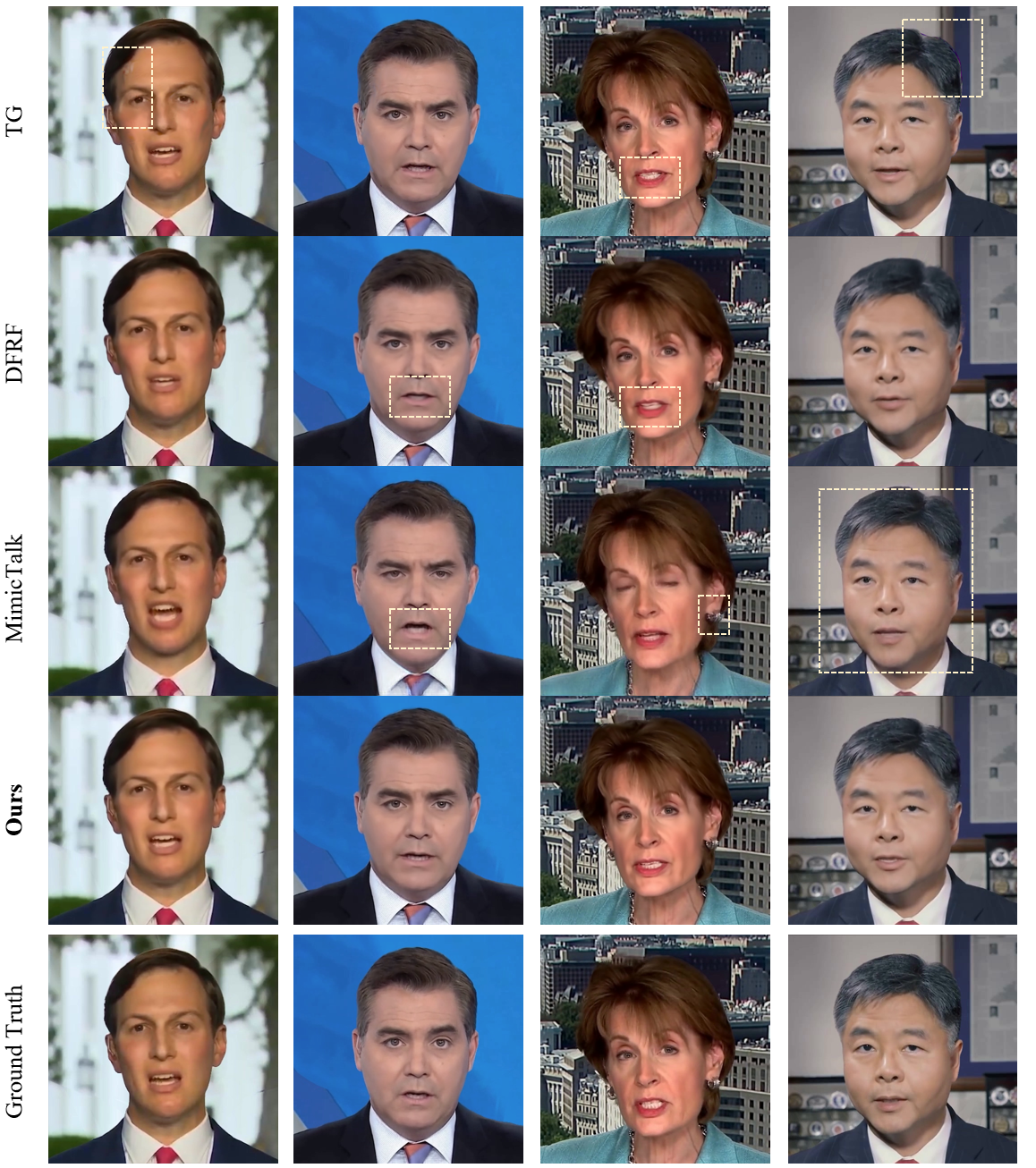}
    \caption{\textbf{High-definition Comparison.} We shot the generated results from models trained with 5s training data. While other methods produce blurry or distorted regions (yellow box), InsTaG successfully reconstructs a faithful appearance for different identities.}
    \label{fig:hd}
    \vspace{1cm}
\end{figure*}

\section{Implementation Details}
\label{sec:impl}

\paragraph{Preprocessing.}
Following most previous radiance-field-based works \cite{guo2021ad, li2023efficient, li2024talkinggaussian}, we assume the head to be with a fixed pose in a canonical space and vary the camera pose to represent head rotation. Specifically, we use a BFM \cite{paysan2009bfm} model to fit the head pose and inversely get the camera pose for each video frame. An off-the-shelf BiSeNet \cite{yu2018bisenet} model is used to segment the head and torso parts and inherit TalkingGaussian \cite{li2024talkinggaussian} to segment the inside mouth and other head areas with EasyPortrait \cite{kvanchiani2023easyportrait}. We represent upper-face expressions with the action units estimated by OpenFace \cite{baltrusaitis2018openface, baltruvsaitis2015openface2}. The audio features are firstly extracted by a pre-trained audio encoder like the DeepSpeech \cite{hannun2014deepspeech} used in the main paper, and then reorganized by a trainable CNN attention network from AD-NeRF \cite{guo2021ad}. 

\paragraph{Model Architecture.}
We build the structure field in 3DGS Talking Head Synthesizer based on the 3DGS official implementation \footnote{https://github.com/graphdeco-inria/gaussian-splatting}. The positional encoder of the Universal Motion Field is built with three 2D multi-resolution hash encoders \cite{muller2022instant}, with a resolution range from 16 to 256 and 64 to 384 separately for the face and inside-mouth branches. Then, a 3-layer MLP decoder is behind with hidden dimensions of 64 and 32 for the two branches. In personalized fields, the positional encoder has the same hyperparameter as in UMF, while the dimension of the MLP decoder decreases to 32 and 16. All the 2D hash encoders are set with 12 levels. More details can be found in the source code.

\paragraph{Optimization.}
Our method inherits the density control strategy of 3DGS to allow the growth of primitives. We start by initializing the radiance field by a random point cloud. In pre-training, we randomly select one frame per iteration from the multi-person data, and calculate the loss to update the model. During adaptation, we first use a small learning rate at the warmup stage with 3000 iterations to build a basic head geometry, and switch on the Motion Aligner in the remaining process. 

\paragraph{Baselines.}
In the main paper, we implement the baselines using their official codes. For the generative models, we use the officially provided weights for reproduction. For all person-specific training, we follow the official instructions to train their models on each video. If available, we uniformly take DeepSpeech \cite{hannun2014deepspeech} as the audio feature extractor for fairness. Specifically, we follow the officially provided scripts to train the model of RAD-NeRF \cite{tang2022rad}, ER-NeRF \cite{li2023efficient}, TalkingGaussian \cite{li2024talkinggaussian} and GaussianTalker \cite{cho2024gaussiantalker} with suggested hyperparameters. For GeneFace \cite{ye2023geneface}, we select the weight of its fine-tuned landmark predictor following the official instruction based on the losses. For DFRF \cite{shen2022dfrf}, we fine-tune the models with 50K iterations, which is a bit larger than the 40K iterations provided in its original paper to ensure convergence. For MimicTalk \cite{ye2024mimictalk}, we use the same training video as the fine-tuning data as well as the style reference, and drive the model by the poses from the ground truth.

\section{Additional Visualization}
\label{sec:vis}
In the supplementary materials, we additionally provide high-definition visualized results to show the advantage of InsTaG in visual quality quantitatively. In Figure \ref{fig:hd}, while baseline methods generate blurry and distorted details (yellow box) with inadequate training data, InsTaG produces high-fidelity talking heads with a faithful appearance. We additionally provide a \textcolor{blue}{supplementary video}, which brings a more intuitive and comprehensive comparison.

\section{Discussions}

\paragraph{Explanation of the design and effect of Negative Contrast Loss.} 
Considering $\mu$ decides the global structure of the fields, while $s$ and $q$ are mainly for fitting texture [26], we only include $\Delta\mu$ in the NCLoss. Once the primitive is deformed with two different $\Delta\mu$, it would represent distinct motions even with the same $s$ and $q$. In our test, adding $s$ and $q$ would not help much.

\paragraph{Why use BFM for head pose estimation? }
BFM is widely used to estimate head pose in many talking head methods, and it has been inherited by all the NeRF and 3DGS-based baselines in our experiment. For fairness, we also adopt it, although some new models like FLAME perform better. We'll advance the use of them in future work.

\paragraph{Non-ideal data inputs.}
Generating high-quality results from non-ideal input is challenging for nearly all methods. By reducing the data demand, our InsTaG makes it easier to capture ideal inputs with a much shorter length. Besides, InsTaG is robust for lower sharpness (first row in Fig. 5) and uneven lighting (two males with side light at video $\sim$08:30). Occlusion or light change may be more challenging. For the former, it may be solved by some in-the-wild techniques. And for ther later, adding external light conditions may help. Confidence can also help to filter the low-quality frames. Another way is to leverage the power of generative models to restore the images to an normal one.

\section{Ethical Consideration}
\label{sec:ethi}
We propose InsTaG and hope our method can promote the healthy development of digital industries. However, several ethical considerations should be noticed.
One key concern pertains to privacy and consent, particularly when training and deploying models on personalized data. To mitigate risks, it is crucial to ensure that all video data in real-world applications should be collected with explicit consent and handled in compliance with data protection regulations. 

Another consideration involves the potential misuse of the technology, such as the creation of deceptive deepfakes or unauthorized impersonation. While generative methods are mostly proposed for positive purposes, their capabilities could be exploited for malicious purposes. To address this, safeguards like embedding traceable digital watermarks or partnering with authentication mechanisms can be integrated to distinguish synthetic from real content. We will share the results to support the deepfake detection community to help people recognize fake content.

\section{Limitation and Future Work}
\label{sec:discuss}
Our proposed InsTaG achieves the fast learning of personalized 3D talking heads from few data, making an advancement in few-shot talking head generation. Nevertheless, some limitations still remain to be addressed in the future.

The first problem is about the extrapolation. Although our pre-training strategy can fetch rich knowledge from existing video data, the basic work principle of the neural fields is still to conduct interpolation between the known signals. Although audio signals can be well generalized by the off-the-shelf audio feature encoder, various upper-face expressions are not easy to be fully covered by the several pre-training data. This would lead to unexpected facial distortion when some unseen expression combinations are input. One of the reasons for this is the instability of the implicit mapping in MLP. We may solve the problem by posing an expression-driving method with a more explicit structure in the future

Second, the adaptation time of InsTaG is possible to be further shortened. Considering the success of some one-shot 3D head methods, the time of structure reconstruction can be skipped if an accurate 3D head is provided from external. Besides, introducing explicit matching rules may help accelerate the alignment between the pre-trained motion field and the new identity head. We will explore more acceleration techniques in the future.

{
    \small
    \bibliographystyle{ieeenat_fullname}
    \bibliography{main}
}



\end{document}